\documentclass[journal]{IEEEtran}
\usepackage[pdftex]{graphicx}
\usepackage{cite}
\usepackage{verbatim}		
\usepackage{amsmath}
\usepackage{amssymb}
\usepackage{amsthm}
\usepackage{mathtools}	
\usepackage{grffile}	
\usepackage[tight,footnotesize]{subfigure}
\usepackage{microtype} 
\usepackage[misc]{ifsym}
\usepackage{xcolor}
\usepackage{url}
\usepackage[ruled,vlined,linesnumbered]{algorithm2e}
\usepackage{bm} 
\usepackage{hyperref}
\hypersetup{
	colorlinks=true,
	linkcolor=blue,
	citecolor=red,
	urlcolor=black,
}

\theoremstyle{remark}
\newtheorem{remarkx}{Remark}
\newenvironment{remark}
{\pushQED{\qed}\remarkx}
{\popQED\endremarkx}

\theoremstyle{definition}
\newtheorem{defn}{Definition}
\newtheorem{assump}{Assumption}
\newtheorem*{problem*}{Problem}
\newtheorem{problem}{Problem}
\newtheorem{feature}{Feature}

\theoremstyle{plain}
\newtheorem{theorem}{Theorem}
\newtheorem{lemma}{Lemma}
\newtheorem{coroll}{Corollary}
\newtheorem{prop}{Proposition}

\newcommand{\dist}{\mathrm{dist}}

\newcommand{\defeq}{:=} 
\newcommand{\matr}[1]{\begin{bmatrix} #1 \end{bmatrix}}
\newcommand{\vmatr}[1]{\begin{vmatrix} #1 \end{vmatrix}}
\newcommand{\transpose}[1]{#1^\top}
\newcommand{\norm}[1]{\left\lVert#1\right\rVert}
\newcommand{\normm}[1]{\big\lVert#1\big\rVert}		

\newcommand{\setp}{\mathcal{P}}
\newcommand{\setc}{\mathcal{C}}
\newcommand{\setm}{\mathcal{M}}
\newcommand{\mbr}[1][{}]{\mathbb{R}^{#1}}	
\newcommand{\diag}[1]{\mathrm{diag}\{ #1\}}
\newcommand{\dw}[1]{\frac{\mathrm{d} #1}{\mathrm{d} w}}
\newcommand{\derivative}[2]{\frac{\mathrm{d} #1}{\mathrm{d} #2}}

\newcommand{\chiup}{\raisebox{2pt}{$\chi$}}
\newcommand{\vf}{\chiup}
\newcommand{\identity}{\mathrm{id}}
\newcommand{\set}[1]{\mathcal{#1}}
\newcommand{\tangentmap}{\bm{\mathrm{D}}}

\newcommand{\proj}[1]{{#1}^{\mathrm{prj}}}
\newcommand{\trs}[1]{{#1}^{\mathrm{trs}}}
\newcommand{\hgh}[1]{{#1}^{\mathrm{hgh}}}
\newcommand{\phy}[1]{{#1}^{\mathrm{phy}}}

\begin{document}
	\title{Singularity-free Guiding Vector Field for \\ Robot Navigation} 
	
	\author{
		Weijia~Yao,  
		H{\' e}ctor~Garcia de Marina, 
		Bohuan~Lin, 
		Ming~Cao
		\thanks{Weijia Yao and Ming Cao are with Institute of Engineering and Technology (ENTEG), Bohuan Lin is with Bernoulli Institute of Mathematics, Computer Science and Artificial Intelligence (BI), all in the University of Groningen, the Netherlands. Hector Garcia de Marina is a Researcher in the Department of Computer Architecture and Automatic Control at the Faculty of Physics, Universidad Complutense de Madrid, 28040 Madrid, Spain.  \texttt{\{w.yao, b.lin, m.cao\}@rug.nl, hgarciad@ucm.es}. }
	}
	
	\maketitle
	
	\begin{abstract}
		In robot navigation tasks, such as UAV highway traffic monitoring, it is important for a mobile robot to follow a specified desired path. However, most of the existing path-following navigation algorithms cannot guarantee global convergence to desired paths or enable following self-intersected desired paths due to the existence of singular points where navigation algorithms return unreliable or even no solutions. One typical example arises in \emph{vector-field guided path-following (VF-PF) navigation algorithms}. These algorithms are based on a vector field, and the singular points are exactly where the vector field diminishes. Conventional VF-PF algorithms generate a vector field of the same dimensions as those of the space where the desired path lives. In this paper, we show that it is mathematically \emph{impossible} for conventional VF-PF algorithms to achieve global convergence to desired paths that are \emph{self-intersected} or even just \emph{simple closed} (precisely, homeomorphic to the unit circle). Motivated by this new impossibility result, we propose a novel method to transform self-intersected or simple closed desired paths to \emph{non-self-intersected} and \emph{unbounded} (precisely, homeomorphic to the real line) counterparts in a \emph{higher-dimensional} space. Corresponding to this new desired path, we construct a singularity-free guiding vector field on a higher-dimensional space. The integral curves of this new guiding vector field is thus exploited to enable global convergence to the higher-dimensional desired path, and therefore the projection of the integral curves on a lower-dimensional subspace converge to the physical (lower-dimensional) desired path. Rigorous theoretical analysis is carried out for the theoretical results using dynamical systems theory. In addition, we show both by theoretical analysis and numerical simulations that our proposed method is an extension combining conventional VF-PF algorithms and trajectory tracking algorithms. Finally, to show the practical value of our proposed approach for complex engineering systems, we conduct outdoor experiments with a fixed-wing airplane in windy environment to follow both 2D and 3D desired paths.
	\end{abstract}
	
	\IEEEpeerreviewmaketitle
	
	\section{Introduction}
	Several robot navigation tasks, such as highway traffic monitoring, underwater pipeline inspection, border patrolling and area coverage, require the fundamental functionality of following a desired path \cite{Siciliano:2007:SHR:1209344}, and new applications are emerging, such as using drones to probe atmospheric phenomena along prescribed paths \cite{lacroix2016fleets}. The path-following navigation problem has attracted the attention from both the robotics community \cite{Goncalves2010,rezende2018robust,michalek2018vfo,hector2017,liang2016combined} and the control community \cite{KAPITANYUK20147342,lakomy2017,michalek2010vector,caharija2016integral,borhaug2008integral,yao2020auto}. In a path-following navigation algorithm, the desired path is usually given in the form of a single connected curve \emph{without temporal information}, and then robots are guided to converge to and move along it with sufficient accuracy. Treating the desired path as a \emph{geometric object} rather than a time-dependent point, path-following navigation algorithms sometimes are able to overcome a number of performance limitations rooted in trajectory tracking \cite{aguiar2008performance}, e.g., inaccuracy due to unstable zero dynamics \cite{seron1999feedback} and difficulty to maintain constant tracking speed \cite{kapitanyuk2017guiding}. Moreover, comparing to trajectory tracking algorithms, there is separate interest for the study of path-following navigation algorithms since they are more suitable for some applications, such as fixed-wing aircraft guidance and control \cite{rezende2018robust,lawrence2008lyapunov,Sujit2014}.
	
	Among different path-following navigation algorithms, those using a \emph{guiding vector field} have been studied widely \cite{Goncalves2010,rezende2018robust,michalek2018vfo,KAPITANYUK20147342,lakomy2017,michalek2010vector}, and we refer to these algorithms as \emph{vector-field guided path-following navigation algorithms (or, VF-PF algorithms)}. The \emph{guidance} feature of the vector field is justified as follows: usually robot \emph{kinematics models} (e.g., single-integrator and double-integrator models \cite{lawrence2008lyapunov,Goncalves2010}) are considered, and the guiding vector field, as its name suggests, provides \emph{guidance} signal inputs to the model. This is valid based on the common assumption that  the \emph{robot-specific} inner-loop \emph{dynamics control} can track these guidance signal inputs effectively \cite{phillips2004mechanics,rubi2019survey,kapitanyuk2017guiding}. Thus, one can simply focus on the guidance layer (i.e., designing a guiding vector field), while considering other control layers separately. Specifically, the guiding vector field is designed such that its integral curves are guaranteed to converge to a predefined geometric desired path. Utilizing the convergence property of the vector field, one can then derive suitable control laws. It is claimed in \cite{Sujit2014} that VF-PF algorithms demonstrate the lowest cross-track error while requiring the least control effort among several other path-following navigation algorithms. In addition, \cite{caharija2015comparison} shows that VF-PF algorithms achieve better path-following accuracy than the integral line-of-sight (ILOS) method \cite{borhaug2008integral}.
	
\begin{figure}
		\centering
		\includegraphics[width=0.5\linewidth]{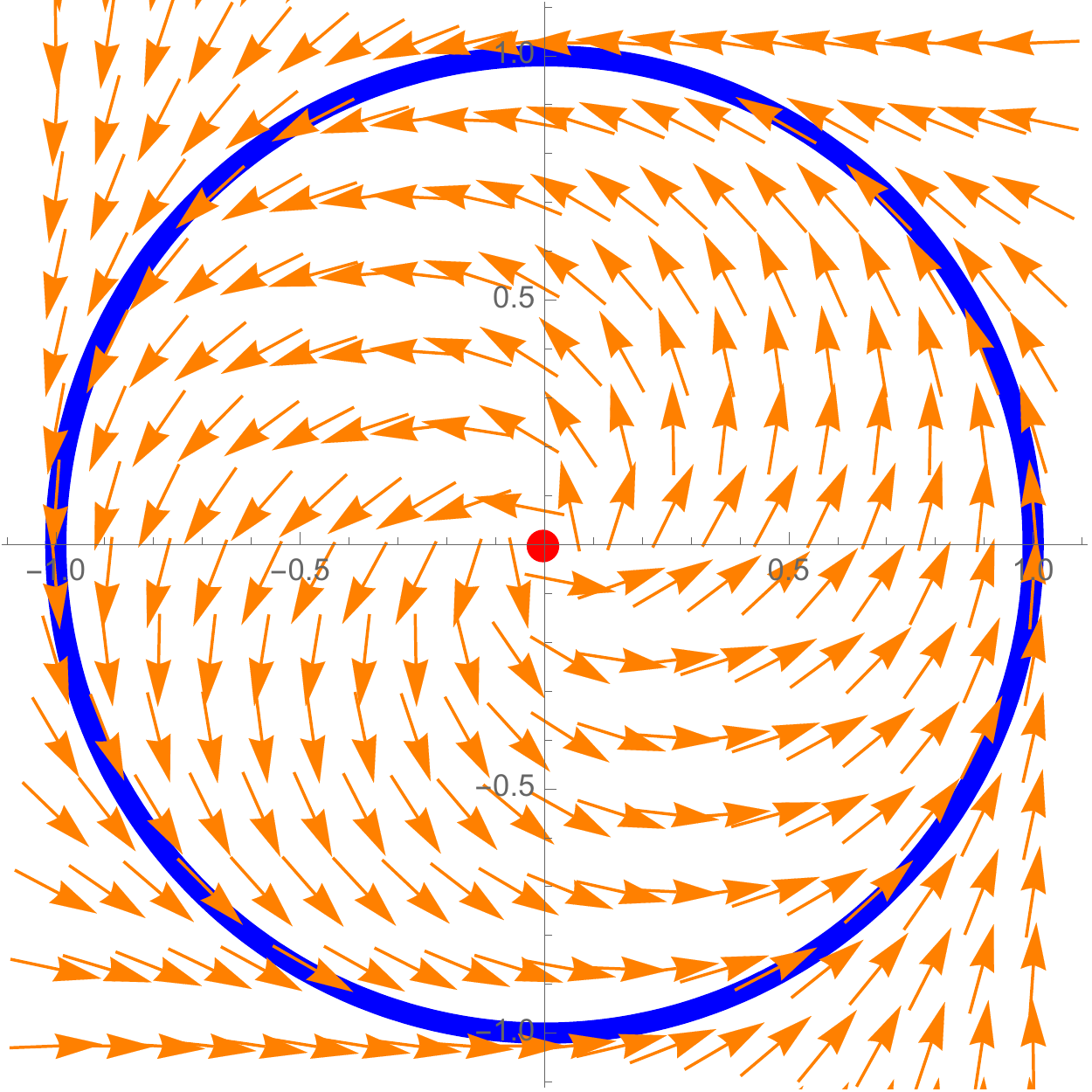}
		\caption{The \emph{normalized} vector field \cite{kapitanyuk2017guiding} for a circle path described by $\phi(x,y) = x^2 + y^2 - 1 = 0$. The red point at the center of the circle is the singular point of the (un-normalized) vector field.}
		\label{fig:circlevf}
	\end{figure}

\begin{figure}
		\centering
		\includegraphics[width=0.5\linewidth]{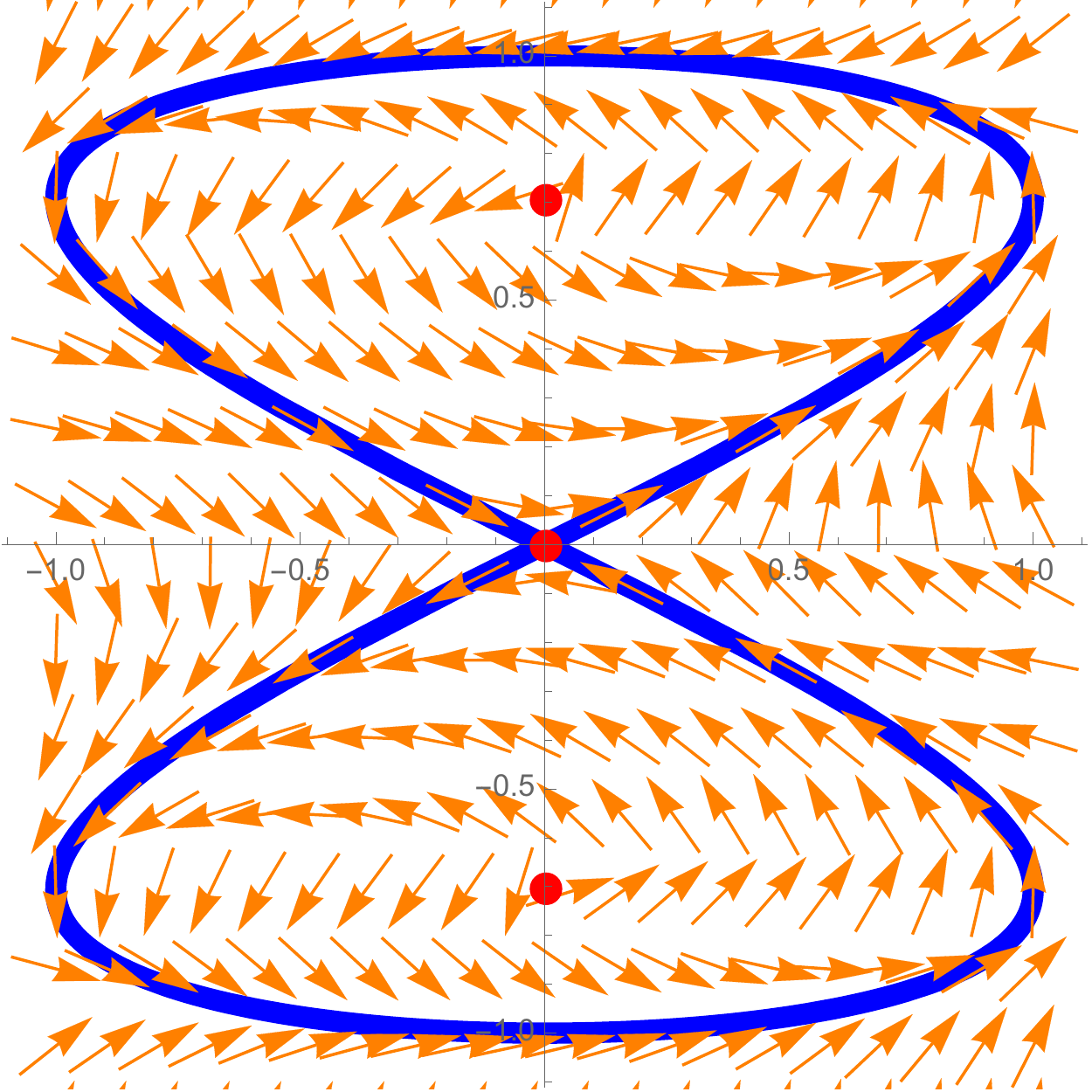}
		\caption{The \emph{normalized} vector field \cite{kapitanyuk2017guiding} for a figure ``8'' path described by $\phi(x,y) = x^2 - 4 y^2 (1-y^2) = 0$. Three red points are the singular points of the (un-normalized) vector field.}
		\label{fig:eightvf}
	\end{figure}

	Although the VF-PF algorithms are intuitive and easy to implement, the rigorous analysis remains nontrivial for general desired paths \cite{kapitanyuk2017guiding,yao2018cdc,de2017guidance,yao2020auto}.  Significant difficulty in the analysis and application of the VF-PF algorithms arises when there are singular points\footnote{A point where a vector field becomes zero is called a \emph{singular point} of the vector field \cite[p. 219]{lee2015introduction}. } in the vector field (see Fig. \ref{fig:circlevf} and Fig. \ref{fig:eightvf}). In such a case, the convergence of trajectories to the desired path cannot be guaranteed globally, and the normalization of the vector field at those points is not well-defined  \cite{kapitanyuk2017guiding,yao2018cdc,Goncalves2010,yao2020auto}. In \cite{Goncalves2010}, it is assumed that these singular points are repulsive to simplify the analysis, while this assumption is dropped in \cite{kapitanyuk2017guiding} for a planar desired path and in \cite{yao2018cdc,yao2020auto} for a desired path in 3D. However, to the best of our knowledge, few efforts have been made on dealing with singular points directly or on eliminating them effectively. Recently, \cite{rezende2018robust} presents a simple treatment of the singular point: the robot does not change its course inside a ball centered at the singular point. Under some conditions, the Lyapunov function evaluated at the exit point is proved to be smaller than that at the entry point. However, these conditions are conservative, since they assume repulsiveness of the singular points, which is not the case, e.g., for saddle points. 
	
	Related to the existence of singular points, one of the challenges for the VF-PF navigation problem is to follow a self-intersected desired path. Many existing VF-PF algorithms (e.g., \cite{Goncalves2010,kapitanyuk2017guiding,nelson2007vector,lawrence2008lyapunov,yao2020auto}) fail to fulfil this task. This is rooted in the fact that the vector field degenerates to zero at the crossing points of a self-intersected desired path, leading to a zero guidance signal, and thus a robot can get stuck on the desired path (see Fig. \ref{fig:eightvf}). Due to the existence of singular points on the desired path, some effective VF-PF algorithms such as \cite{kapitanyuk2017guiding,yao2018cdc,yao2020auto} become invalid simply because the assumptions are violated in this case. In fact, this task is also challenging for other existing path-following methods, since in the vicinity of the crossing points, many methods are ``ill-defined''. For example, the line-of-sight (LOS) method \cite{fossen2003line} is not applicable as there is not a unique projection point in the vicinity of the intersection of the desired path. Indeed, many existing path-following navigation algorithms either focus on simple desired paths such as circles or straight lines \cite{Sujit2014,Zhu2013,nelson2007vector}, or only deal with desired paths that are sufficiently smooth \cite{kapitanyuk2017guiding,yao2018cdc,Goncalves2010,yao2020auto}. One might retreat to the virtual-target path-following navigation algorithm \cite{soetanto2003adaptive}. In this method, a virtual target has its own dynamics travelling on the desired path; thus the path-following navigation problem is implicitly converted to a target tracking problem. Although through this conversion, it is possible for a robot to follow a self-intersected desired path, this method is inherently a tracking approach, and thus may inherit the performance limitations mentioned before, such as limited path-following accuracy.


	Another challenging task arising from the VF-PF methods is the description of the desired path, which is crucial for the derivation of the vector field. For generality, the desired path is usually determined by the intersection of several (hyper)surfaces represented by the zero-level sets of some implicit functions \cite{Goncalves2010,kapitanyuk2017guiding,yao2018cdc,rezende2018robust,sgorbissa20103D,yao2020auto}. For planar desired paths, for example, the implicit function of a star curve might be as complicated as that in \cite{liang2016combined}, while for desired paths in a higher-dimensional space, it is counter-intuitive to create (hyper)surfaces such that the intersection is precisely the desired path, such as a helix. On the other hand, many geometric curves are described by parametric equations \cite{do2016differential} rather than implicit functions. It is possible to transform the parametric equations to implicit functions and then derive the vector field, but this might not always be feasible and is computationally expensive. The restrictive characterization of the desired path limits the applicability of VF-PF algorithms to some extent.
	
	In this paper, we improve the VF-PF methodology in the sense that we address the aforementioned three issues: the existence of singular points, the obstacle of dealing with self-intersected paths, and the difficulty of representing a generic desired path. Specifically, based on the design of guiding vector fields in \cite{yao2018cdc}, we use an intuitive idea to eliminate singular points of the vector field so that global convergence to the desired path, even if self-intersected, is guaranteed. The general idea is to extend the dimensions of the vector field and eliminate singular points simultaneously. This procedure naturally leads to a simple way to transform the descriptions of desired paths from parameterized forms to the intersection of several (hyper)surfaces, which are required in creating a guiding vector field.  
	
	It is important to clarify the terminology used throughout this paper. In many VF-PF algorithms, the desired path is a geometric object which is not necessarily parameterized. In a precise mathematical language, we assume that the desired path is a \emph{one-dimensional connected differential manifold} \cite{yao2020auto}. Therefore, we can generally classify desired paths into two categories: those homeomorphic to the unit circle $\mathbb{S}^1$ if they are compact, and those homeomorphic to the real line $\mbr[]$ otherwise \cite[Theorem 5.27]{lee2010topologicalmanifolds}. This assumption is not a restriction, since many desired paths in practice, such as a circle, an ellipse, a Cassini oval, a spline and a straight line, satisfy this assumption. For ease of exposition, we refer to those desired paths homeomorphic to the unit circle as \emph{simple closed} desired paths, and those homeomorphic to the real line as \emph{unbounded} desired paths. Note that \emph{self-intersected} paths do not satisfy this assumption. Nevertheless, we will introduce in the sequel how to \emph{transform} a self-intersected physical desired path to a non-self-intersected and unbounded higher-dimensional desired path such that the assumption holds to apply our algorithm. 
	
	Here, we summarize the major \textbf{contributions} of our paper:

	Firstly, we show by rigorous topological analysis that guiding vector fields with the same dimension as the desired path (e.g., \cite{yao2018cdc,kapitanyuk2017guiding,liang2016combined,michalek2018vfo}) cannot guarantee the global convergence to a simple closed or self-intersected desired path (see Theorem \ref{note_thm2} in Section \ref{sec_impos}). With the dichotomy of convergence discussed in the paper, this implies that singular points of the vector field \emph{always exist} for a \emph{simple closed} or \emph{self-intersected} desired path regardless of which hypersurfaces one uses to characterize the desired path. This explains why many vector-field guided path-following navigation algorithms in the literature cannot guarantee global convergence in the Euclidean space to a simple closed desired path. We note that excluding singular points is important in practice (e.g., for fixed-wing aircraft guidance and navigation) since degenerated or pathological solutions of system dynamics can be safely avoided. Therefore, this topological obstacle is the primary motivation of the subsequent theoretical development including the introduction of extended dynamics (see Section \ref{sec_extended}) and the creation of singularity-free guiding vector fields (see Section \ref{sec_newgvf}).
	
	Secondly, due to the aforementioned topological obstruction, we improve the existing VF-PF algorithms such that \textbf{all} singular points are removed, and global convergence of trajectories to the desired path is rigorously guaranteed (see Section \ref{sec_extended} and Section \ref{sec_newgvf}). We overcome this topological obstruction by changing the topology of the desired paths. Specifically, we transform a physical \emph{simple closed} or \emph{self-intersected} desired path to a new \emph{unbounded} and \emph{non-self-intersected} desired path in a \emph{higher-dimensional space}. We then derive the corresponding guiding vector field on this higher-dimensional space, which is guaranteed to have \emph{no} singular points. 
	
	Thirdly, our proposed method to create this new singularity-free guiding vector field is proved to enjoy several appealing features (see Subsection \ref{subsec_feature}). For example, we provide theoretical guarantees for global exponential convergence of trajectories of system dynamics to the desired path. In addition, the new system dynamics with the singularity-free guiding vector field is robust against perturbation, such as noisy position measurements (see Feature \ref{feature3} in Subsection \ref{subsec_feature}). Moreover, using our proposed method, it becomes straightforward to represent (hyper)surfaces of which the intersection is the new higher-dimensional desired path, as long as a parametrization of the physical (lower-dimensional) desired path is available (see Feature \ref{feature2} in Subsection \ref{subsec_feature}).   
	
	Last but not least, we successfully conduct experiments using a fixed-wing aircraft to verify the effectiveness of our proposed VF-PF algorithm in 3D (see Section \ref{sec_experiment}), in addition to the experiment with an e-puck robot \cite{mondada2009puck} in our previous preliminary work \cite{yao2020mobile2}. This verifies the practical significance of our proposed method for highly complex autonomous vehicles. We also discuss and conclude that our proposed VF-PF algorithm combines and extends features of the conventional VF-PF algorithms and trajectory tracking algorithms (see Section \ref{sec_disc}). While we do not claim that our proposed new singularity-free guiding vector field is always superior than traditional trajectory tracking algorithms in every application scenario (such as quadcopter attitude tracking), we emphasize that it significantly improves conventional VF-PF algorithms by providing a global solution and enabling the path-following behavior of complicated or unconventional desired paths (e.g., a self-intersected Lissajous curve). This is imperative and irreplaceable in applications such as fixed-wing aircraft guidance and navigation where convergence to and propagation along a desired path from \emph{every} initial position is required.

	This paper is a significant extension of our preliminary version in \cite{yao2020mobile2} in the following aspects: 1)  \cite{yao2020mobile2} only considers physical planar desired paths and the \emph{higher-dimensional} 3D vector field, but we generalize all the results to $n$-dimensional; 2) \cite{yao2020mobile2} only investigates a \emph{linear transformation operator}, but we introduce a more general \emph{transformation operator} which can be linear or nonlinear in this paper. Compared with \cite{yao2020mobile2}, we present five new results: 1) we prove a topological theorem regarding the feasibility of global convergence to desired paths, which serves as the primary motivation for the subsequent development; 2) we provide rigorous guarantees to justify the features of our proposed approach; 3) we provide experiments with a fixed-wing aircraft in 3D to verify the practicality of the proposed approach for complex engineering systems; 4) we elaborate on some important aspects in the implementation of the proposed approach; 5) we provide new insightful discussion of our proposed approach and existing algorithms. Although our previous work in \cite{yao2020auto} introduces part of the theoretical foundation for this paper, it does not solve the singularity problem that will be addressed here.
	
	The remainder of this paper is organized as follows. Section \ref{sec_vf} introduces conventional guiding vector fields for path following. In Section \ref{sec_impos}, a theorem about the impossibility of global convergence to simple closed or self-intersected desired paths using the conventional VF-PF navigation algorithm is elaborated. This is the main motivation for the design of higher-dimensional guiding vector fields, which will be utilized in Section \ref{sec_extended} through \emph{extended dynamics}. Based on the previous sections, the construction approach of \emph{singularity-free} guiding vector fields is presented in Section \ref{sec_newgvf}. In addition, several appealing features of this method are highlighted in this section. Then experiments with a fixed-wing aircraft are conducted to validate the theoretical results in Section \ref{sec_experiment}. Following this, Section \ref{sec_disc} discusses how our proposed approach can be viewed as a combined extension of VF-PF algorithms and trajectory tracking algorithms.  Finally, Section \ref{sec_conclu} concludes the paper and indicates future work. Some theoretical proofs are provided in the Appendix.
	
	We present here some notations and basic concepts that are used throughout the paper.
	
	\textbf{Notations}:  Given a positive integer $n$, the \emph{distance} between a point $p_0 \in \mathbb{R}^n$ and a nonempty set $\mathcal{S} \subseteq \mathbb{R}^n$ is denoted by $ \dist(p_0, \mathcal{S}) = \dist(\mathcal{S}, p_0 ) \defeq  \inf\{ || p - p_0|| : p \in \mathcal{S} \}$. The distance between two nonempty sets $\mathcal{A}$ and $\mathcal{B}$ is $\dist(\mathcal{A}, \mathcal{B}) = \dist(\mathcal{B}, \mathcal{A}) = \inf\{ ||a-b||: a \in \mathcal{A}, b \in \mathcal{B} \}$. If $\xi$ is a differentiable function of time $t$, then the derivative of $\xi$ with respect to time is $\dot{\xi}(t)$. The notation $\norm{\cdot}$ denotes the Euclidean norm of a vector or the matrix two-norm of a matrix. The normalization of a vector $v$ is denoted by $\hat{v}$; namely, $\hat{v}=v / \norm{v}$. The transpose of a vector $v$ is denoted by $\transpose{v}$. Suppose there is a function $\rho: \mathcal{C} \to \mathcal{D}$ from a nonempty set $\mathcal{C}$ to a nonempty set $\mathcal{D}$. The image of a subset $\mathcal{F} \subseteq \mathcal{C}$ under $\rho$ is the subset $\rho(\mathcal{F}) \subseteq \mathcal{D}$ defined by $\rho(\mathcal{F}) \defeq \{ \rho(x)  \in \mathcal{D} : x \in \mathcal{F}\}$. Given two functions $f: X \to Y, g: Y \to Z$, $g \circ f$ denotes the composition of these two functions. The notation $\defeq$ means ``defined to be''.
	
	\textbf{Basic concepts:} A trajectory $\xi: [0, +\infty) \to \mathbb{R}^n$ asymptotically converges to a nonempty set $\mathcal{B} \subseteq \mathbb{R}^n$ if for any $\epsilon>0$, there exists $T>0$ such that $\dist(\xi(t),\mathcal{B}) < \epsilon$ for all $t > T$. If the trajectory can only be maximally prolonged to $t_*<\infty$ \cite[Chapter 3]{khalil2002nonlinear}, then we say that it converges to the set $\mathcal{B}$ as $t$ approaches $t_*$, if for any $\epsilon>0$, there exists $\delta>0$ such that $ \dist(\xi(t),\mathcal{B})<\epsilon$ for $|t-t_*| < \delta$. Two subsets $U \subseteq \mbr[m]$ and $V \subseteq \mbr[k]$ are \emph{homeomorphic}, denoted by $U \approx V$, if there exists a continuous bijection $\varphi: U \to V$ of which the inverse is also continuous. The map $\varphi$ is called a \emph{homeomorphism} \cite{lee2010topologicalmanifolds}.
	
	\section{Guiding Vector Fields for Path Following} \label{sec_vf}
	In this section, we introduce the vector-field guided path-following (VF-PF) navigation problem and guiding vector fields. We formally define the VF-PF navigation problem as follows.
	\begin{problem}[VF-PF navigation problem] \label{def0}
		Given a desired path $\setp \subseteq \mbr[n]$, the VF-PF navigation problem is to design a continuously differentiable vector field $\vf: \mbr[n] \to \mbr[n]$ for the differential equation $\dot{\xi}(t) = \vf \big( \xi(t) \big)$ such that the two conditions below are satisfied: 
		\begin{enumerate}
			\item There exists a neighborhood $\mathcal{D} \subseteq \mbr[n]$ of the desired path $\setp$ in \eqref{eqpath2d} such that for all initial conditions $\xi(0) \in \mathcal{D}$, the distance $\dist(\xi(t), \setp)$ between the trajectory $\xi(t)$ and the desired path $\setp$ approaches zero as time $t \to \infty$; that is, $\lim_{t \to \infty} \dist(\xi(t), \setp) = 0$;
			\item If a trajectory starts from the desired path, then the trajectory stays on the path for $t \ge 0$ (i.e., $\xi(0) \in \setp \implies \xi(t) \in \setp$ for all $t \ge 0$). In addition, the vector field on the desired path is nonzero (i.e., $0 \notin \vf(\setp)$).
		\end{enumerate}  
	\end{problem}
	In this paper, we only investigate the guiding vector field in the Euclidean space $\mbr[n]$, but it can be extended to a general smooth manifold \cite{yao2021topo}. For easy understanding of the guiding vector field on $\mbr[n]$ in the sequel, we first introduce the one defined on the two-dimensional Euclidean space $\mbr[2]$.
	
	\subsection{Preliminaries on 2D VF-PF Control} \label{sec_pre}
	In \cite{kapitanyuk2017guiding}, the desired path $\setp$ is described by the zero-level set of an implicit function:
	\begin{equation} \label{eqpath2d}
	\setp = \{ (x,y) \in \mbr[2] : \phi(x,y) = 0\},
	\end{equation} 
	where $\phi: \mbr[2] \to \mbr[]$ is twice continuously differentiable. In this description, $\setp$ is a subset of $\mbr[2]$. The description is different from some other works where the desired path is a parameterized differentiable curve (e.g., \cite{aguiar2008performance}); that is, a differentiable map $f: I \to \mbr[n]$ of an open interval $I=(a,b)$ of the real line $\mbr[]$ into $\mbr[n]$ \cite{do2016differential}. From the definition, we observe that the mathematical object in \eqref{eqpath2d} is actually the \emph{trace} of a parameterized curve $f$ \cite{do2016differential}, or the image of a mapping $f$. However, this description of the desired path without any parametrization is common in the field of VF-PF control \cite{yao2018cdc,Goncalves2010,rezende2018robust,michalek2018vfo,consolini2010path,morro2011path,do2015global,chen2011curve,yao2020auto}. One of the advantages is that the vector field can be derived directly from the function $\phi(\cdot)$, independent of the specific parametrization of the desired path. Another advantage is that we can replace the calculation of the Euclidean distance\footnote{This calculation is generally difficult since one needs to find the closest point on the desired path to $\xi$; e.g., it is not trivial for even an ellipse.} $\dist(\xi, \setp)$ between a point $\xi \in \mbr[2]$ and the desired path $\setp$ simply by the value of $|\phi(\xi)|$. For simplicity, rather than referring to $\setp$ in \eqref{eqpath2d} as ``the trace of a parameterized curve'', we call $\setp$ the desired path throughout the paper. In fact, one feature of the VF-PF navigation problem is that the desired path $\setp$ is described by a \emph{one-dimensional connected submanifold}, so we have the extra freedom of choosing different parametrizations or analytic expressions for the same desired path. 
	
	If the desired path is non-self-intersected, then a valid 2D vector field $\vf: \mbr[2] \to \mbr[2]$ to solve the VF-PF navigation problem is \cite{kapitanyuk2017guiding}:
	\begin{equation} \label{eqvf}
	\vf(x,y) = E \nabla \phi(x,y) - k \psi \big( \phi(x,y) \big) \nabla \phi(x,y), 
	\end{equation}
	where $E\in SO(2)$ is the $90^\circ$ rotation matrix\footnote{In fact, the matrix is $-E$ in \cite{kapitanyuk2017guiding}, but we use $E$ for conventional simplicity. This only affects the direction of the motion (forward or backward)  on the desired path.} $\left[\begin{smallmatrix}0 & -1 \\ 1 & 0\end{smallmatrix}\right]$, and $\psi: \mbr[] \to \mbr[]$ is a strictly increasing function satisfying $\psi(0)=0$. For simplicity, one can choose $\psi \big( \phi(x,y) \big) = \phi(x,y)$. The first term of the vector field is ``tangential'' to the desired path, thus enables a robot to move along the desired path, while the second term of the vector field is perpendicular to the first term, helping the robot move closer to the desired path. Therefore, intuitively, the vector field guides the robot to move towards and along the desired path at the same time. 
	
	After the preliminaries on the guiding vector field on $\mbr[2]$ (i.e., 2D vector field), we are ready to introduce in the next subsection the more abstract guiding vector field on $\mbr[n]$, where $n \ge 2$. 

	\subsection{General Guiding Vector Field} \label{sec_gvf}
	Based on the description of a desired path $\setp \subseteq \mbr[n]$, the corresponding guiding vector field for path following is derived in this subsection. Some mild assumptions along with motivation and justification are presented. 
	
	\subsubsection{Guiding Vector Field Structure}
	We introduce the general method of designing a vector field $\vf: \mbr[n] \to \mbr[n]$ corresponding to a desired path $\setp \subseteq \mbr[n]$, where $n \ge 2$. The vector field is a generalization of those in \cite{kapitanyuk2017guiding,yao2018cdc}, while its structure is the same as that in \cite{Goncalves2010}. However, we present assumptions that are crucial for this method and abandon the unnecessary assumption in \cite{Goncalves2010} about the repulsiveness of the set of singular points.
	
	Suppose a desired path in the $n$-dimensional Euclidean space is described by the intersection of $(n-1)$ hypersurfaces; that is, 
	\begin{equation} \label{path1}
	\mathcal{P} = \{ \xi \in \mathbb{R}^n : \phi_i(\xi)=0, \; i=1,\dots, n-1\},
	\end{equation}
	where $\phi_i:\mathbb{R}^n \to \mathbb{R}$, $i=1, \dots, n-1$, are of differentiability class $C^2$. It is naturally assumed that $\setp$ in \eqref{path1} is \emph{nonempty} and \emph{connected}. We further require the regularity of the desired path as shown later in Assumption \ref{assump1}. For better understanding, $\phi_i(\cdot)=0$ can be regarded as $(n-1)$ constraints, resulting in a one degree-of-freedom desired path. 
	
	\begin{remark} \label{remark_dimension}
		Topologically, the desired path $\setp$ itself is one-dimensional, independent of the dimensions of the Euclidean space where it lives. However, with slight abuse of terminology and for convenience, the desired path $\setp$ is called an \emph{$n$-D (or $n$D) desired path} if it lives in the $n$-dimensional Euclidean space $\mbr[n]$ and \textbf{not} in any lower-dimensional subspace $\mathcal{W} \subseteq \mbr[n]$ (i.e., the \emph{smallest subspace} the desired path lives in). For example, a planar desired path might be defined in the three-dimensional Euclidean space $\mbr[3]$, but we only consider the two-dimensional subspace $\mathcal{W} \subseteq \mbr[2]$ where it is contained, and it is thus natural to call it a $2$D (or $2$-D) desired path rather than a $3$D (or $3$-D) desired path. Sometimes, for simplicity, we refer to a tangent vector field on the $n$-dimensional Euclidean space $\mbr[n]$ as an \emph{$n$-dimensional vector field}, and we say that \emph{this vector field is $n$-dimensional}.
	\end{remark}
	
	The vector field $\vf:\mathbb{R}^n \to \mathbb{R}^n$ is designed as below:
	\begin{equation} \label{gvf}
	\vf = \times(\nabla \phi_1, \dots, \nabla \phi_{n-1}) - \sum_{i=1}^{n-1} k_i \phi_i \nabla \phi_i,
	\end{equation}
	where $\nabla \phi_i$ is the gradient of $\phi_i$, $k_i>0$ are constant gains, and $\times: \mathbb{R}^n \times \dots \times  \mathbb{R}^n \to \mathbb{R}^n$ is the generalized cross product. In particular, let $p_i=\transpose{(p_{i1} , \cdots , p_{in})} \in \mathbb{R}^n$, $i=1,\dots,n-1$, and $\bm{b_j} \in \mathbb{R}^n$ be the standard basis column vector with the $j$th component being $1$ and the other components being $0$. Then
	\begin{align*}
	\times(p_1, \dots, p_{n-1}) =& \vmatr{p_{12} & \cdots & p_{1n} \\ \vdots & \vdots & \vdots \\ p_{n-1,2} & \cdots & p_{n-1,n}} \bm{b_1} + \dots + \\ &(-1)^{n-1} \vmatr{p_{11} & \cdots & p_{1,n-1} \\ \vdots & \vdots & \vdots \\ p_{n-1,1} & \cdots & p_{n-1,n-1}} \bm{b_n}.
	\end{align*}
	An intuitive \emph{formal expression} is
	\[
	\times(p_1, \dots, p_{n-1}) =  \vmatr{ \bm{b_1} & \bm{b_2} & \cdots & \bm{b_n} \\ p_{11} & p_{12} & \cdots & p_{1n} \\ \vdots & \vdots & \vdots & \vdots \\ p_{n-1,1} & p_{n-1,2} & \cdots & p_{n-1,n}}.
	\]
	In other words, first forming an $(n-1)$ by $n$ matrix by putting vectors $\transpose{p_1}, \dots, \transpose{p_{n-1}}$ row by row, and then the $k$th component of $\times(p_1, \dots, p_{n-1})$ is obtained as $(-1)^{k-1}$ multiplying the determinant of the submatrix deleting the $k$th column. This is similar to the wedge product in \cite{Goncalves2010}. The generalized cross product has the following property \cite[Proposition 7.2.1]{galbis2012vector}:
	\begin{lemma}[Orthogonality] \label{lemma1}
		It holds that $\times(p_1, \dots, p_{n-1})$ is orthogonal to each of the vectors $p_1, \dots, p_{n-1}$. 
	\end{lemma}
	\begin{remark} \label{remark1}
		Due to Lemma \ref{lemma1}, the physical interpretation of the vector field in \eqref{gvf} is clear. The first term $\times(\nabla \phi_1, \dots, \nabla \phi_{n-1})$, being perpendicular to each $\nabla \phi_i$, provides a tangential direction to each surfaces $\phi_i(\xi)=0$, and hence ``pushes'' the robot \textbf{along} the desired path. The forward or backward direction of movement along the desired path is determined by the order of the gradient vectors. Thus if the motion needs to be reversed, it is sufficient to swap any two of these vectors. We call this term the \emph{propagation term}. The latter term $-\sum_{i=1}^{n-1} k_i \phi_i \nabla \phi_i$ provides a direction \textbf{towards} those surfaces, acting as a ``pulling force'' to the desired path. We call this term the \emph{converging term}. 
	\end{remark}
	
	To simplify the notations, we define a matrix $N(\xi) = \big( \nabla \phi_1(\xi) , \cdots , \nabla \phi_{n-1}(\xi) \big) \in \mathbb{R}^{n \times (n-1)}$, a positive definite gain matrix $K = \diag{k_1, \dots, k_{n-1}} \in \mathbb{R}^{(n-1) \times (n-1)}$ and a $C^2$ function $e: \mbr[n] \to \mbr[n-1]$ by stacking $\phi_i$; that is,
	\begin{equation} \label{eqe}
	e(\xi) = \transpose{ \big( \phi_1(\xi) , \cdots , \phi_{n-1}(\xi) \big)} \in \mathbb{R}^{n-1}.
	\end{equation}
	In addition, we define $\nabla_{\times} \phi : \mbr[n] \to \mbr[n]$ by $ \xi \in \mbr[n] \mapsto \times \big( \nabla \phi_1(\xi), \dots, \nabla \phi_{n-1}(\xi) \big)$. Therefore, the vector field \eqref{gvf} can be compactly written as
	\begin{equation} \label{eqvf2}
	\vf(\xi) = \nabla_{\times} \phi(\xi) - N(\xi) K e(\xi).
	\end{equation}
	Using this notation, the desired path is equivalent to 
	\begin{equation} \label{path2}
	\mathcal{P} = \{ \xi \in \mathbb{R}^n : e(\xi)=0\}.
	\end{equation}
	We call $e(\xi)$ the \emph{path-following error} or simply \emph{error} between the point $\xi \in \mbr[n]$ and the desired path $\setp$. An intuitive example is a 2D circle described by the zero-level sets of $e(x,y)=\phi_1(x,y)=x^2+y^2-r^2$. As the point $(x,y)$ approaches the circle, the norm of the path-following error $\norm{e(x,y)}=|x^2+y^2-r^2|$ decreases. When $\norm{e(x,y)}=0$, the point is right on the path. The use of $\norm{e(\cdot)}$ is more convenient than that of the distance function $\dist(\cdot, \setp)$. However, there are subtle differences between the norm of the path-following error $\norm{e(\cdot)}$ and the distance $\dist(\cdot, \setp)$. For example, when the norm of the path-following error converges to zero, the trajectory may diverge from the desired path (see Example $3$ in \cite{yao2018cdc}). To avoid this pathological situation, assumptions are proposed in the next subsection.
	
	\begin{remark} \label{remark_variants}
		The vector field in \eqref{gvf} is basic while effective in VF-PF navigation problems. Firstly, we note that many existing studies only deal with simple desired paths such as a circle or a straight line (or a combination of them)  \cite{Sujit2014,caharija2016integral,nelson2007vector}, but the vector field in \eqref{gvf} is designed for any \emph{generic} sufficiently smooth desired path. Secondly, many vector fields in the literature can be seen as \emph{variants} of the vector field in \eqref{gvf}. One type of variants are generated by adding $\phi_i$-dependent gains to the converging term or (and) the propagation term in \eqref{gvf} \cite{michalek2018vfo,kapitanyuk2017guiding,liang2016combined}. Another type of variants adds time-varying gains or an additional time-varying component \cite{Goncalves2010,lawrence2008lyapunov}. Thus, the vector fields in \cite{michalek2018vfo,kapitanyuk2017guiding,liang2016combined,yao2019integrated} can be regarded as 2D variants of \eqref{gvf}, and those in \cite{lakomy2017,yao2018cdc,yao2020auto,kapitanyuk2017guiding2,liang2016combined} as 3D	variants of \eqref{gvf}. Therefore, the study of the basic vector field in \eqref{gvf} is of great significance. Note that we do \textbf{not} consider time-varying gains or components in the vector field as \cite{Goncalves2010,lawrence2008lyapunov} do. For one thing, this simplifies the structure of the vector field and facilitates the practical implementation; for another, this clarifies the topological property of these vector fields as studied in Section \ref{sec_impos}. For convenience, we refer to these (time-invariant) vector fields in the literature as \emph{conventional vector fields}.
	\end{remark}
	
\subsubsection{Assumptions} \label{subsecb}
	To justify using the norm of the path-following error $\norm{e(\cdot)}$ instead of $\dist(\cdot, \setp)$, we need some assumptions that are easily satisfied in practice. These assumptions are based on \cite{yao2018cdc,kapitanyuk2017guiding}, but are extended to $\mbr[n]$. To this end, we define two sets. The \emph{singular set} consisting of singular points of a vector field is defined as below:
	\begin{equation} \label{singularset}
	\setc = \{ \xi \in \mbr[n] : \vf(\xi)=0 \}.
	\end{equation}
	Another related set is 
	\begin{equation} \label{eqsetm}
	\setm = \left\{ \xi \in \mbr[n] : N(\xi) K e(\xi) = 0 \right\}.
	\end{equation}
	It can be proved that $\setm = \setp \cup \setc$. 
	\begin{lemma} \label{lemma_mpc}
		It holds that $\setm = \setp \cup \setc$.
	\end{lemma}
	\begin{proof}
	See Appendix \ref{appA1}.
	\end{proof}
	Now we are ready to propose the following assumptions.	
	\begin{assump} \label{assump1}
		There are no singular points on the desired path. More precisely, $\dist(\mathcal{C}, \mathcal{P}) > 0$.
	\end{assump}
	\begin{assump} \label{assump2}
		In view of \eqref{path2}, as the norm of the path-following error $\norm{e(\xi)}$ approaches zero, the trajectory $\xi(t)$ approaches the desired path $\setp$. Similarly, in view of \eqref{eqsetm}, as the ``error'' $\norm{N(\xi) K e(\xi)}$ approaches zero, the trajectory $\xi(t)$ approaches the set $\setm$.
	\end{assump}
	
	Assumption \ref{assump1} leads to the following statement about the topological property of the desired path, which is an extension of \cite{yao2018cdc}. 
	\begin{lemma} \label{lemmanifold}
		The zero vector $\bm{0} \in \mbr[n-1]$ is a regular value of the $C^2$ function $e$ in \eqref{eqe}, and hence the desired path $\mathcal{P}$ is a $C^2$ embedded submanifold\footnote{See \cite[pp. 105, 98-99]{lee2015introduction} for the definitions of a regular value and an embedded submanifold.} in $\mathbb{R}^n$. 
	\end{lemma}
	\begin{proof}
	See Appendix \ref{appA2}
	\end{proof}
	\begin{remark} \label{remark5}
		Henceforth the ``regularity'' of the desired path is guaranteed; namely, the desired path $\mathcal{P}$ is assumed to be a one-dimensional connected manifold, which can generally be classified into those homeomorphic to the unit circle if they are compact, and those homeomorphic the real line otherwise \cite[Theorem 5.27]{lee2010topologicalmanifolds}. Thus throughout the paper, we use the notions of \emph{simple closed desired paths} and \emph{desired paths homeomorphic to the unit circle} interchangeably. The same applies to \emph{unbounded desired paths} and \emph{desired paths homeomorphic to the real line}.	Note that self-intersected desired paths do not satisfy Assumption \ref{assump1}, as shown later in Proposition \ref{prop1}, but we will propose a method in Section \ref{sec_newgvf} to transform them into unbounded and non-self-intersected desired paths, which are then homeomorphic to the real line $\mbr[]$.
	\end{remark}
	
	Assumption \ref{assump2} is satisfied for many practical cases such as many polynomial functions and trigonometric functions; examples are demonstrated in \cite{kapitanyuk2017guiding,yao2018cdc,Goncalves2010,rezende2018robust}. Since the same desired path can be characterized by various choices of $\phi_i(\cdot)$, these assumptions are crucial to exclude some pathological cases \cite[Example 3]{yao2018cdc}. The precise mathematical formulation of Assumption \ref{assump2} is presented in Appendix \ref{appA3}. 
	
	\section{Issues on the Global Convergence to Desired Paths} \label{sec_impos}
	In this section, we show that, under some conditions, it is not possible to guarantee global convergence to desired paths using the existing VF-PF algorithms as introduced in Section \ref{sec_vf}. More specifically, given a desired path $\setp \subseteq \mbr[n]$ as described in \eqref{path1}, we investigate solutions (trajectories) of the autonomous ordinary differential equation:
	\begin{equation} \label{eqode}
	\dot{\xi}(t) = \vf \big( \xi(t) \big),
	\end{equation}
	where $\vf$ is defined in \eqref{eqvf2}. We consider the cases of self-intersected desired paths and simple closed desired paths respectively as follows.
	
	We first show that the crossing points of a self-intersected desired path $\setp$ are singular points of the corresponding vector field $\vf$ in \eqref{gvf}. 
	\begin{prop} \label{prop1}
		If the desired path $\setp$ in \eqref{path1} is self-intersected, then the crossing points of the desired path are singular points of the vector field $\vf$ in \eqref{gvf}.
	\end{prop}
	\begin{proof}
	See Appendix \ref{appB1}
	\end{proof}
	\begin{remark}
	This proposition shows that $0 \in \vf(\setp)$ when $\setp$ is a self-intersected desired path, and therefore, the VF-PF navigation problem (Problem \ref{prop1}) cannot be addressed as the second requirement about $0 \notin \vf(\setp)$ is always violated. In fact, note that Assumption \ref{assump1} does not hold in this case, but we will propose in the sequel an approach to transform a self-intersected desired path such that Assumption \ref{assump1} holds and the VF-PF problem is solved.
	\end{remark}
	In Fig. \ref{fig:eightvf}, for example, the 2D desired path resembling the figure ``$8$'' is self-intersected. It can be numerically calculated that the vector field at the crossing point is zero. This is intuitive in the sense that there is no ``preference'' for the vector at this point to point to either the left or right portion of the desired path, leaving the only option of zero.
	
	Now we consider simple closed desired paths. In the planar case, due to the Poinc{\'a}re-Bendixson theorem \cite[Corollary 2.1]{khalil2002nonlinear}, there is at least one singular point of the \emph{2D} vector field in the region enclosed by the simple closed desired path. Thus we can conclude that global convergence to a simple closed \emph{planar} desired path is not possible. However, this conclusion cannot be trivially generalized to the higher-dimensional case since the Poinc{\'a}re-Bendixson theorem is restricted to the planar case. Nevertheless, we can still reach this conclusion with some topological analysis.
	
	\begin{prop} \label{note_prop2new}
		If an $n$-D desired path $\mathcal{P} \subseteq \mbr[n]$ described by \eqref{path1} is simple closed, under the dynamics \eqref{eqode} where the guiding vector field $\vf: \mbr[n] \to \mbr[n]$ is in \eqref{gvf}, then it is not possible to guarantee the global convergence of trajectories of \eqref{eqode} to the desired path $\mathcal{P}$; precisely, the domain of attraction of $\setp$ cannot be $\mbr[n]$.
	\end{prop}
	\begin{proof}
		See Appendix \ref{appB2}.
	\end{proof}
	
	Based on Proposition \ref{prop1} and Proposition \ref{note_prop2new}, we can reach the following key statement about the impossibility of global convergence to some desired paths.
	\begin{theorem}[Impossibility of global convergence] \label{note_thm2}
		If an $n$-D desired path $\mathcal{P} \subseteq \mbr[n]$ described by \eqref{path1} is \emph{simple closed} or \emph{self-intersected}, then it is not possible to guarantee the global convergence to the desired path with respect to the dynamics in \eqref{eqode} with the $n$-dimensional guiding vector field $\vf$ in \eqref{gvf}; more precisely, the domain of attraction of $\setp$ cannot be $\mbr[n]$.
	\end{theorem}
	\begin{proof}
		If the desired path $\mathcal{P}$ is self-intersected, then by Proposition \ref{prop1}, there is at least one singular point on the desired path. Obviously, the path-following problem formulated by Problem \ref{def0} cannot be solved. If the desired path $\mathcal{P}$ is simple closed, then the global convergence to the desired path is impossible by Proposition \ref{prob2}.
	\end{proof}
	\begin{remark}
		We note that the topological obstacle to global convergence to the desired path roots in two aspects: i) the geometry of the desired path: being either simple closed or self-intersected; ii) the time-invariance property of the vector field. Although we show this topological obstacle only for the basic vector field in \eqref{gvf}, this obstacle also exists for other variants of vector fields as listed in Remark \ref{remark_variants}. This is because the two aspects that lead to the topological obstruction mentioned above also hold for these time-invariant vector fields. Note that a state-dependent positive scaling (e.g., the normalization) of vector fields does not affect the topological properties of interest (i.e., the phase portrait, or the convergent results) \cite[Proposition 1.14]{chicone2006ordinary}.
	\end{remark}
	To overcome this topological obstacle and satisfy Assumption \ref{assump1} even for self-intersected desired paths, we propose a new idea in the sequel to construct unbounded and non-self-intersected desired paths from the originally simple closed or self-intersected desired paths by ``stretching'' them in a higher-dimensional space. Indeed, such a higher-dimensional desired path will codify or contain information about the (lower-dimensional) physical desired path. Based on the proposed higher-dimensional desired paths, we can derive a guiding vector field on this higher-dimensional space and show that its singular set is empty. However, to take advantage of the new guiding vector field, we need to \emph{transform} (or \emph{project} in the linear transformation case) its integral curves into a lower-dimensional subspace that contains the information of the physical desired path. The details of transformation into another space will be discussed in Section \ref{sec_extended}, and the detailed construction of a singularity-free guiding vector field on a higher-dimensional space will be presented in Section \ref{sec_newgvf}.
	
	\section{Extended Dynamics and Convergence Results} \label{sec_extended}
	In this section and the subsequent sections, we consider an $m$-dimensional Euclidean space $\mbr[m]$, where $m>n$. The reason is self-evident as the paper develops, but it is not necessary to bother with this difference now. To proceed, we introduce the \emph{extended dynamics} and derive related convergence results. The extended dynamics relates to a \emph{transformation operator}, which is define as follows:
	\begin{defn}[Transformation operator]
		A transformation operator is a function $G_l : \mbr[m] \to \mbr[m]$ which is twice continuously differentiable and \emph{globally Lipschitz continuous} with the Lipschitz constant $l$.
	\end{defn} 
	One can observe that the corresponding Jacobian matrix function of a transformation operator $\tangentmap G_l = \partial G_l / \partial x : \mbr[m] \to \mbr[m \times m]$ is locally Lipschitz continuous, where $x$ is the argument of $G_l$.	The transformation operator is able to transform a space into another space (or subspace). One example is a \emph{linear transformation operator} defined by $G_l(x)=Ax$, where $A$ is a nonzero matrix, called the \emph{matrix representation}\cite[Remark 6.1.15]{tao2015analysis2} of this particular linear transformation operator $G_l$. Having defined the transformation operator, we can now introduce the extended dynamics as follows.
	\begin{lemma}[Extended dynamics] \label{lemma2}
		Let $\vf: \mathcal{D} \subseteq \mbr[m] \to \mbr[m]$ be a vector field that is locally Lipschitz continuous. Given an initial condition $\xi(0)=\xi_0 \in \mathcal{D}$, suppose that $\xi(t)$ is the unique solution to the differential equation $\dot{\xi}(t) =\vf \big( \xi(t) \big)$, then $\big( \xi(t), \trs{\xi}(t) \big) \in \mbr[2m]$, where $\trs{\xi}(t) \defeq G_l \big( \xi(t) \big)$ and $G_l$ is a transformation operator, is the unique solution to the following initial value problem:
		\begin{equation} \label{eq6}
		\begin{cases}
		\dot{\xi}(t) =\vf \big(\xi(t) \big) & \xi(0) = \xi_0\\
		\trs{\dot{\xi}}(t) = \tangentmap G_l \big( \xi(t) \big) \cdot \vf \big( \xi(t) \big) & \trs{\xi}(0) = G_l(\xi_0).
		\end{cases}
		\end{equation}
		Moreover, if the trajectory $\xi(t)$ asymptotically converges to some set $\mathcal{A} \ne \emptyset \subseteq \mbr[m]$, then $\trs{\xi}(t)$ asymptotically converges to the \emph{transformed set}
		$
		\trs{\mathcal{A}} \defeq G_l(\mathcal{A}) = \{ p \in \mbr[m] :  p = G_l(q), \; q \in \mathcal{A} \}.
		$
	\end{lemma}
	\begin{proof}
		See Appendix \ref{appC1}.
	\end{proof}
	
	We call the ordinary differential equation with the initial condition in \eqref{eq6} \emph{extended dynamics}. Correspondingly, we call $\trs{\xi}(t) \defeq G_l \big( \xi(t) \big)$ the \emph{transformed solution} or \emph{transformed trajectory} of \eqref{eq6}. Before presenting the next corollary related to the VF-PF navigation problem, we first define the \emph{transformed desired path} and \emph{transformed singular set}.
	\begin{defn}
		The \emph{transformed desired path} $\trs{\setp}$ of $\setp \subseteq \mbr[m]$ in \eqref{path2} and the \emph{transformed singular set} $\trs{\setc}$ of $\setc \subseteq \mbr[m]$ in \eqref{singularset} are defined below:
		\begin{align}
		\trs{\setp} &\defeq G_l(\setp) = \{ p \in \mbr[m] : p = G_l(q), \; q \in \setp \} \label{pp} \\
		\trs{\setc} &\defeq G_l(\setc) = \{ p \in \mbr[m] : p = G_l(q), \; q \in \setc \}. \label{pc}
		\end{align}
	\end{defn}
	
	In some practical applications, it is desirable to scale the vector field to have a specified constant length. This is useful if a robot takes the vector field as the control input directly and is required to move at a constant speed. In this case, the properties of the integral curves of the scaled vector field are stated in the corollary below. Recall that the solution $x(t)$ to an initial value problem $\dot{x}=f(x), \; x(0)=x_0$, where $f(x)$ is sufficiently smooth, is not always possible to be prolonged to infinity. In other words, the solution might only  be well-defined in a finite time interval $[0,t^*)$, where $t^* < \infty$ \cite{chicone2006ordinary}. The time instant $t^*$ is called the maximal prolonged time of the solution.
	
	\begin{coroll} \label{corol_prj_dyn}
		Suppose the desired path $\setp$ in \eqref{path2} is unbounded (i.e., $\setp \approx \mbr[]$). Let $\vf: \mathcal{D} \subseteq \mbr[m] \to \mbr[m]$ be the vector field defined in \eqref{gvf}. Suppose $\xi(t)$ is the unique solution to the initial value problem $\dot{\xi}(t) = s \hat{\vf} \big( \xi(t) \big), \; \xi(0)=\xi_0 \notin \mathcal{C}$, where $s>0$ is a constant and $\hat{\cdot}$ is the normalization operator. Consider the following dynamics
		\begin{equation} \label{eq7}
		\begin{cases}
		\dot{\xi}(t) = s \hat{\vf} \big( \xi(t) \big) & \xi(0) = \xi_0 \notin \setc \\
		\trs{\dot{\xi}}(t) = \tangentmap G_l \cdot s \hat{\vf} \big( \xi(t) \big) & \trs{\xi}(0) = G_l(\xi_0),
		\end{cases}
		\end{equation}
		where $G_l$ is a transformation operator. Suppose $t^* \le \infty$ is the maximal prolonged time of the transformed solution $\trs{\xi}(t)$ to \eqref{eq7}. Then $\trs{\xi}(t)$ asymptotically converges to the transformed desired path $\trs{\setp}$ in \eqref{pp} as $t \rightarrow \infty$ or the transformed singular set $\trs{\setc}$ in \eqref{pc} as $t \rightarrow t^*$.
	\end{coroll}
	\begin{proof}
		See Appendix \ref{appC2}.
	\end{proof}
	\begin{remark}
		Due to the normalization of the vector field in \eqref{eq7}, the right-hand side of the differential equation is not well defined at singular points of the vector field. Therefore, if the transformed singular set $\trs{\setc}$ is bounded, then the maximal interval to which the transformed trajectory $\trs{\xi}(t)$ can be prolonged is only finite when the transformed trajectory $\trs{\xi}(t)$ is converging to $\trs{\setc}$. This happens when the initial value $\xi(0)$ is in the invariant manifold of the singular set $\setc$. 
	\end{remark}
	
	The previous lemma states that the \emph{transformed trajectory} converges to either the \emph{transformed desired path} or the \emph{transformed singular set} for initial conditions $\trs{\xi}(0) \in \mbr[m] \setminus G_l(\setc)$, while the latter case is undesirable. A preferable situation is where the (transformed) singular set is empty. Moreover, as indicated by Theorem \ref{note_thm2}, to seek for global convergence, the only possibility is to consider unbounded and non-self-intersected desired paths (i.e., $\setp \approx \mbr[]$). Therefore, we reach the following corollary.
	
	\begin{coroll}[Global convergence to $\trs{\setp}$] \label{coroll_global_conv}
		Suppose the desired path $\setp$ in \eqref{pp} is unbounded (i.e., $\setp \approx \mbr[]$). If $\setc=\emptyset$ (equivalently, $\trs{\setc}=\emptyset$), then the transformed trajectory $\trs{\xi}(t)$ of \eqref{eq7} globally asymptotically converges to the transformed desired path $\trs{\setp}$ as $t \rightarrow \infty$ in the sense that the initial condition $\xi(0)$ (and hence $\trs{\xi}(0)$) can be arbitrarily chosen in $\mbr[m]$.
	\end{coroll}
	As will be shown later, only the second differential equation of \eqref{eq6} or \eqref{eq7} is relevant to the physical robotic system. This corollary thus motivates us to design a (higher-dimensional) vector field such that the singular set is empty, in which case global convergence to the (transformed) desired path is guaranteed. In the next section, we will introduce an intuitive idea to ``stretch'' a possibly simple closed or self-intersected physical desired path and create a higher-dimensional singularity-free vector field.
	
	\section{High-dimensional Singularity-free Guiding Vector Field Construction} \label{sec_newgvf}
	Following Section \ref{sec_impos} and Section \ref{sec_extended}, in this section, we explain how to implicitly construct an unbounded desired path from the physical desired path (possibly simple closed or self-intersected) together with a higher-dimensional guiding vector field without any singular points (a.k.a, singularity-free guiding vector field). 
	
	For simplicity, we restrict the transformation operator $G_l: \mbr[m] \to \mbr[m]$ to a linear one defined by $G_l(x) = P_a x$, where $P_a \in \mbr[m \times m]$ is a nonzero matrix defined by
	\begin{equation} \label{linear_prj}
	P_a = I - \hat{a} \transpose{\hat{a}},
	\end{equation}
	where $I$ is the identity matrix of suitable dimensions and $\hat{a}= a / \norm{a} \in \mbr[m]$ is a normalized nonzero vector. In this case, $G_l$ is actually a linear transformation that projects an arbitrary vector to the hyperplane orthogonal to the given nonzero vector $a$, and $P_a$ is called the \emph{matrix representation} of $G_l$ \cite{tao2015analysis2}. One can observe that the linear transformation $G_l$ is globally Lipschitz continuous with the Lipschitz constant $l=\norm{P_a}=1$, where $\norm{\cdot}$ is the induced matrix two-norm. In addition, the Jacobian is simply $\tangentmap G_l=P_a$. 
	
	Before formulating the problem in the sequel, we define the \emph{coordinate projection function} $\pi_{(1,\dots,n)}: \mbr[m] \to \mbr[n]$ as 
	\[
		\pi_{(1,\dots,n)}(x_1,\dots,x_n,\dots,x_m) = (x_1, \dots, x_n),
	\]
	where $m>n$. In other words, the coordinate projection function $\pi_{(1,\dots,n)}$ takes only the first $n$ components of an $m$-dimensional vector and generates a lower-dimensional one. 
	
	\begin{problem}  \label{prob2}
		Given an $n$-D physical desired path\footnote{Recall the notion of an $n$-D desired path in Remark \ref{remark_dimension}.} $\phy{\setp}$ in $\mbr[n]$, we aim to find an $m$-D desired path $\hgh{\setp}$ in $\mbr[m]$, where $m > n$, which satisfies the following conditions: 
		\begin{enumerate}
			\item there exist functions $\phi_i(\cdot)$, $i=1,\dots,m-1$, such that $\hgh{\setp}$ is described by \eqref{path1};
			\item the singular set $\hgh{\setc}$ of the \emph{higher-dimensional} vector field $\hgh{\vf}:\mbr[m] \to \mbr[m]$ in \eqref{gvf} corresponding to $\hgh{\setp}$ is empty;
			\item there exists a transformation operator $G_l: \mbr[m] \to \mbr[m]$ such that $\pi_{(1,\dots,n)}(\trs{\setp}) = \phy{\setp}$, where the transformed desired path $\trs{\setp}= G_l(\hgh{\setp})$.
		\end{enumerate}
	\end{problem}
	\begin{remark}
		It is important to distinguish among the \emph{physical desired path} $\phy{\setp}$, the \emph{higher-dimensional desired path} $\hgh{\setp}$ and the \emph{transformed desired path} $\trs{\setp}$. A major difference is the dimensions of their ambient space; that is, $\phy{\setp} \subseteq \mbr[n]$, while $\hgh{\setp}, \trs{\setp} \subseteq \mbr[m]$ and $m>n$. Although the higher-dimensional desired path $\hgh{\setp}$ and the transformed desired path $\trs{\setp}$ both live in $\mbr[m]$, the transformed desired path $\trs{\setp}$ lives in a subspace $\set{W} \subseteq \mbr[m]$ \emph{probably} with $\dim(\set{W}) < m$ since $\trs{\setp}=G_l(\hgh{\setp})$. Indeed, for the case of a linear transformation operator in \eqref{linear_prj}, the transformed desired path $\trs{\setp}=P_a(\hgh{\setp})$ lives in the orthogonal complement subspace $\set{W}$ of the linear space spanned by the vector $a$ (i.e., $\mathrm{span}\{a\}$), and $\dim(\set{W})=m-1 < m$.
	\end{remark}
	
	Next, we propose the solution to Problem \ref{prob2} in Subsection \ref{subsec_construct}. Having found the higher-dimensional desired path $\hgh{\setp}$, then we can directly derive the corresponding vector field $\hgh{\vf}$ defined on $\mbr[m]$ by \eqref{gvf}. Some features of the approach illustrated in Subsection \ref{subsec_construct} are highlighted in Subsection \ref{subsec_feature}.
	
	\subsection{Construction of a Singularity-free Guiding Vector Field} \label{subsec_construct}
	Suppose an $n$-D physical desired path $\phy{\setp}$ is parameterized by
	\begin{equation} \label{eq8}
	x_1 = f_1(w), \, \dots,\,  x_{n} = f_{n}(w),
	\end{equation}
	where $w \in \mbr[]$ is the parameter of the desired path and $f_i \in C^2, \, i=1,\dots,n$. We can simply let 
	\begin{equation} \label{eqsurface1}
	\phi_1(\xi) = x_1 - f_1(w), \; \dots, \; \phi_{n}(\xi) = x_{n} - f_{n}(w),
	\end{equation}
	where $\xi=(x_1, \dots, x_{n}, w)$ has an additional coordinate $w$ now and is an $m$-dimensional vector, where $m=n+1$. So the $m$-D desired path is 
	\begin{equation} \label{eq_high_p}
	\resizebox{0.99\hsize}{!}{$
		\hgh{\setp}=\{ \xi=(x_1, \dots, x_{n}, w) \in \mbr[m] : \phi_i(\xi)=0,\, i=1,\dots, n \}.
		$}
	\end{equation}
	Thus the first requirement of Problem \ref{prob2} is met. Intuitively, the new higher-dimensional desired path $\hgh{\setp}$ is obtained by ``stretching'' the $n$-D desired path $\phy{\setp}$ along the additional virtual $w$-axis. From the higher-dimensional desired path $\hgh{\setp} \subseteq \mbr[m]$ in \eqref{eq_high_p}, we obtain the corresponding guiding vector field on the higher-dimensional space $\mbr[m]$  by \eqref{gvf}:
	\[
		\hgh{\vf} = \nabla_{\times} \phi - \sum_{i=1}^{n} k_i \phi_i \nabla\phi_i.
	\]
	It can be calculated that $\nabla\phi_i = \transpose{ \big(0, \dots, 1, \dots, -f'_i(w) \big)}$ for $i=1,\dots,n$, where $f'_i(w) \defeq \dw{f_i(w)}$ and $1$ is the $i$-th component of the gradient vector. Therefore, 
	\[
	\nabla_{\times}\phi = \times (\nabla \phi_1, \dots, \nabla \phi_{n})  = (-1)^{n} \matr{f'_1(w) \\ \vdots \\ f'_n(w) \\ 1} \in \mbr[m] = \mbr[n+1].
	\]
	It is interesting that the $m$-th coordinate of this vector is a constant $(-1)^{n}$ regardless of the specific parametric form of the desired path. This means that $\norm{\nabla_{\times} \phi(\xi)} \ne 0$ for $\xi \in \mathbb{R}^{m}$ globally. From Lemma \ref{lemma1}, we know that the propagation term $\nabla_{\times} \phi$ of the vector field is always linearly independent from the converging term $ \sum_{i=1}^{n} k_i \phi_i \nabla\phi_i$ unless they are zero vectors. However, as we have shown that $\norm{\nabla_{\times} \phi} \ne 0$ in $\mathbb{R}^{m}$ globally, this reveals the appealing property that the vector field $\hgh{\vf}(\xi) \ne 0$ for any point $\xi \in \mathbb{R}^{m}$, implying that there are no singular points in the higher-dimensional space $\mbr[m]$; i.e., $\hgh{\setc} = \emptyset$. Thus, the second requirement of Problem \ref{prob2} (as well as a related condition in Corollary \ref{coroll_global_conv}) is satisfied. 
	
	To let the third requirement of Problem \ref{prob2} be satisfied, we retreat to a linear transformation operator with a matrix representation $P_a$. One of the simplest linear transformation operators corresponds to $a=\bm{b_{n+1}} \in \mbr[m]$, which is a standard basis column vector with the $(n+1)$-th component being $1$ and the other components being $0$. This can be used to transform an $m$-dimensional space to an $n$-dimensional subspace by ``zeroing'' the last coordinate. Specifically, we let $a=\bm{b_{n+1}}$, then the matrix representation of the linear transformation operator is $P_a=\left[\begin{smallmatrix}I_{n \times n} & \bm{0} \\ \bm{0} & 0\end{smallmatrix}\right]$, where $\bm{0}$ are zero vectors that are of suitable dimensions. Observe that the $n$-D desired path $\phy{\setp} \subseteq \mbr[n]$ is the orthogonal projection of the higher-dimensional desired path $\hgh{\setp} \subseteq \mbr[m]$ on the plane where $w=0$; that is,
	\[
	\pi_{(1,\dots,n)}(\hgh{\setp})=\pi_{(1,\dots,n)}(\trs{\setp})=\phy{\setp}.
	\]
	Therefore, the third requirement of Problem \ref{prob2} is also satisfied. By the construction in \eqref{eq_high_p}, the higher-dimensional desired path $\hgh{\setp} \subseteq \mbr[m]$ satisfying all the conditions in Problem \ref{prob2} is thus found. Ultimately, we can take advantage of the new ``well-behaved'' guiding vector field $\hgh{\vf} \in \mbr[m]$ derived from $\hgh{\setp} \subseteq \mbr[m]$ as mentioned above. This result is formally stated in the following theorem.
	\begin{theorem} \label{thm2}
		Suppose an $n$-D physical desired path $\phy{\setp} \subseteq \mbr[n]$ is parameterized by \eqref{eq8}. If $\phi_1, \dots, \phi_n$ are chosen as in \eqref{eqsurface1}, then there are no singular points in the corresponding guiding vector field $\hgh{\vf}: \mbr[n+1] \to \mbr[n+1]$ defined on the $(n+1)$-dimensional space $\mbr[n+1]$. Let $a=\bm{b_{n+1}}$ for the linear transformation operator $P_a$. Suppose the transformed trajectory of the \emph{extended dynamics} \eqref{eq7} is $\trs{\xi}(t) \defeq \transpose{\big( x_1(t), \dots, x_{n}(t), w(t) \big)}$. Then the \emph{projected transformed trajectory}
		\[
			\proj{\xi}(t) \defeq \pi_{(1,\dots,n)} \big( \trs{\xi}(t) \big) =  \transpose{\big( x_1(t), \dots, x_{n}(t) \big)}
		\]
		globally asymptotically converges to the physical desired path $\phy{\setp}$ as $t \to \infty$.
	\end{theorem}
	\begin{proof}
		By \eqref{gvf} and \eqref{eqsurface1}, the guiding vector field on the $(n+1)$-dimensional space $\mbr[n+1]$ is 
		\begin{equation} \label{gvf_high}
			\hgh{\vf}(x_1, \dots, x_{n}, w) = \matr{(-1)^{n}  f_1'(w) - k_1 \phi_1  \\ \vdots \\ (-1)^{n}  f_{n}'(w) - k_{n} \phi_{n} \\ (-1)^{n} + \sum_{i=1}^{n} k_i \phi_i f_i'(w) }.
		\end{equation}
		As discussed before, the singular set $\hgh{\setc} = \emptyset$. According to Corollary \ref{coroll_global_conv}, $\trs{\xi}(t)$ globally asymptotically converges to the transformed desired path $\trs{\setp} = G_l(\hgh{\setp}) = P_a(\hgh{\setp})$ as $t \rightarrow \infty$. Since $\transpose{a} \trs{\xi} = \transpose{a} P_a \xi = 0$, the $(n+1)$-th coordinate $w(t)$ of the transformed trajectory $\trs{\xi}(t)$ is equal to $0$, meaning that the transformed trajectory $\trs{\xi}(t)$ lies on the subspace $\mathcal{W} \defeq \{(x_1,\dots, x_{n+1}) \in \mbr[n] : x_{n+1} =0 \}$. Therefore, the projected transformed trajectory $\proj{\xi}(t)=\pi_{(1,\dots,n)}\big( \trs{\xi}(t) \big)$ globally asymptotically converges to the physical desired path $\phy{\setp}$.
	\end{proof}
	\begin{remark}
		Note that the proof of convergence to the physical desired path $\phy{\setp}$ is indirect. The norm of the path-following error $\norm{e(\cdot)}=\norm{\big( \phi_1(\cdot), \dots, \phi_n(\cdot) \big)}$ captures the distance to the higher-dimensional desired path $\hgh{\setp}$, taking into account the additional coordinate $w$ as well. It is shown first that in the higher-dimensional space $\mbr[n+1]$, the norm of the path-following error $\norm{e(\cdot)}$ approaches zero asymptotically. Then the convergence to the transformed desired path $\trs{\setp}$ is obtained from Corollary \ref{corol_prj_dyn} (or Corollary \ref{coroll_global_conv}). Due to the special choice of the linear transformation operator $P_a$, where $a = \bm{b_{n+1}}$, the transformed desired path $\trs{\setp}$ is ``almost'' the same as the physical desired path $\phy{\setp}$, except that it has an additional but constant coordinate $w(t) \equiv 0$. 
	\end{remark}
	
	We have shown that, by extending the vector field from $\mbr[n]$ to $\mbr[n+1]$, the new guiding vector field does not have any singular points. Therefore, by using the extended dynamics, the convergence to the physical desired path is guaranteed globally. When $n>3$, this case corresponds to some configuration spaces, such as the robot arm joint space in a smooth manifold embedded in $\mbr[n]$. See \cite{yao2021topo} for more details.
	
	\subsection{Features of the Approach} \label{subsec_feature}
	There are several intriguing features of our proposed approach discussed above in Section \ref{subsec_construct}. These features are summarized below, and the corresponding theoretical guarantees are presented in Appendix \ref{app2}. For ease of narration and without loss of generality, we take the case of a 2D physical desired path $\phy{\setp} \subseteq \mbr[2]$ for discussion (i.e., $n=2$).
	\begin{feature} \label{feature1}
		The corresponding higher-dimensional desired path $\hgh{\setp}=\{ \xi \in \mbr[2+1] : \phi_1(\xi)=0, \phi_2(\xi)=0 \}$ is not self-intersected. This is due to the fact that a crossing point must be a singular point (see Proposition \ref{prop1}), but we have shown that there are no singular points in the higher-dimensional guiding vector field. In fact, the parameter of the desired path $w$ in \eqref{eq8} is implicitly transformed to an additional coordinate of the higher-dimensional desired path. Thus the physical planar desired path $\phy{\setp}$ is ``cut'' and ``stretched'' into the three-dimensional Euclidean space, and becomes unbounded and non-self-intersected along the additional dimension. The significance of this feature is that even a self-intersected physical desired path $\phy{\setp}$ described by \eqref{eq8} can be successfully followed by using the new singularity-free guiding vector field, which in fact corresponds to a non-self-intersected ``stretched'' desired path $\hgh{\setp}$. 
	\end{feature}
	\begin{feature} \label{feature2}
		This approach facilitates the expression of hypersurfaces characterized by implicit functions $\phi_i$. Usually, a parameterized form of the desired path is more readily available than the hypersurfaces of which the intersection is the desired path. Therefore, given the parameterized form in \eqref{eq8}, we do not need to convert them into $\phi(x,y)=0$ and derive the corresponding 2D vector field. Instead, by simply defining two $\phi$ functions as in \eqref{eqsurface1}, we obtain a singularity-free vector field $\hgh{\vf}$ defined on $\mbr[2+1]$.
	\end{feature}
	\begin{feature} \label{feature3}
		One only needs to examine the boundedness of $|f'_i(z)|$, $i=1,2$, in the vicinity of the higher-dimensional desired path $\hgh{\setp}$ to guarantee both the property of local exponential vanishing of the norm of the path-following error $\norm{e}$ and the property of robustness against disturbance of the system dynamics \eqref{eqode}, while these properties usually require more conditions to be satisfied for general vector fields \cite{yao2020auto}. The theoretical guarantees are shown in Proposition \ref{prop_expo} and Proposition \ref{prop_iss} in Appendix \ref{app2} respectively.
	\end{feature}
	\begin{feature} \label{feature4}
		Only Assumption \ref{assump2} is required. Since the new guiding vector field does not have any singular points, the other assumption, Assumption \ref{assump1}, is vacuously true. This is independent of the specific parametrizations of the desired path in \eqref{eq8}. The intuitive Assumption \ref{assump2} holds for many practical examples; thus one might safely ignore it in practice \cite{yao2020auto,yao2018cdc,kapitanyuk2017guiding}.
	\end{feature}
	\begin{feature}
		The additional virtual coordinate can be used to realize scalable distributed multi-robot coordinated path-following navigation by adding a consensus term and guarantee collision avoidance by using a safety barrier certificate \cite{yao2021multi}.
	\end{feature}

\section{Experiments with an autonomous aircraft} \label{sec_experiment}

In this section, we will demonstrate the effectiveness of our path tracking approach with an autonomous fixed-wing aircraft. In particular, we are going to verify the tracking of both 2D and 3D self-intersected desired paths. All the related software has been developed within the open-source project for autopilots \emph{Paparazzi} \cite{gati2013open,de2017guidance,hector2017}. In fact, the codes are ready for others to implement their trajectories and only the corresponding parametric equations are needed\footnote{\url{https://github.com/noether/paparazzi/tree/gvf\_advanced/sw/airborne/modules/guidance/gvf\_parametric}.}.

\subsection{The autonomous aircraft and airfield}
\begin{figure}
	\centering
	\includegraphics[width=0.8\columnwidth]{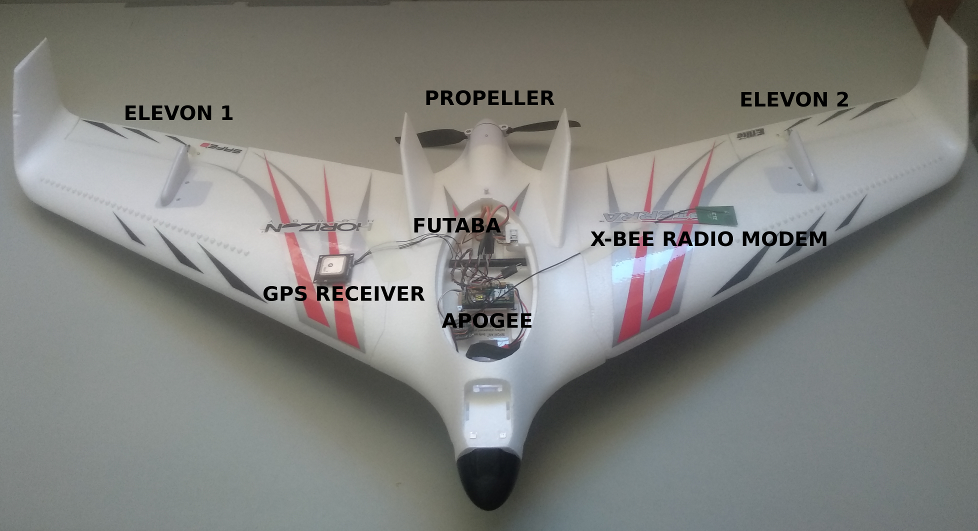}
	\caption{Autonomous Opterra 1.2m equipped with Paparazzi's Apogee autopilot. The airframe is built by E-Flite / Horizon Hobby company.}
	\label{fig: opterra}
\end{figure}

For the experiments, we use one off-the-shelf 1.2m delta-wing Opterra as shown in Fig. \ref{fig: opterra}. Two \emph{elevons} actuate the aircraft at the wings and one motor acts in pushing the configuration. The vehicle's electronics consists of the autopilot \emph{Apogee}\footnote{\url{https://wiki.paparazziuav.org/wiki/Apogee/v1.00}.}, an Ublox GPS receptor, a Futaba receiver, and a X-Bee S1 radio modem. The Apogee's core is an STM32F4 microcontroller where our algorithm runs with a fixed frequency of 50Hz, and all the relevant data are logged in an SD card at 100Hz. The ground segment consists of a standard laptop with another X-Bee S1 radio modem to monitor the telemetry and a Futaba transmitter in case of taking over manual control of the vehicle.

The flights took place on July 18, 2020, in Ciudad Real (Spain) with GPS coordinates $(39.184535, -4.020797)$ degrees. The weather forecast reported 36 Celsius degrees and a South wind of 14 km/h.

\subsection{Aircraft's guidance system design}
We employ a decoupled vertical and horizontal model for setting the aircraft's guiding reference signals. In particular, accounting for the non-holonomic lateral constraint of the aircraft, we consider the following unicycle model
	\begin{equation} \label{model}
		\dot{x} = v \cos \theta \quad \dot{y} = v \sin \theta \quad \dot{\theta} = u_\theta \quad \dot{z} = u_z,
	\end{equation}
	where $(x,y,z)$ is the 3D position, $\theta$ is the heading angle on the XY plane, $v$ is the ground speed, $u_\theta$ is the angular velocity control/guiding signal to change the heading, and $u_z$ is the guiding signal for the climbing velocity. In this section, we are going to show how to design the guiding signals $u_\theta$ and $u_z$, which are injected into the control system of the aircraft that deals with the non-trivial couplings of the lateral and vertical modes. In particular, $u_\theta$ is tracked by banking the aircraft depending on the current speed $v$ and \emph{pitch} angle to achieve a \emph{coordinated turn}\footnote{\url{https://en.wikipedia.org/wiki/Coordinated_flight}.}, and $u_z$ is tracked by controlling the pitch angle and the propulsion to vary the lift and the vertical component of the pushed force coming from the propeller\footnote{We leave the reader to check the details of the employed low-level controllers at \url{http://wiki.paparazziuav.org/wiki/Control_Loops}.}. The experimental results will show the compatibility of our path tracking algorithm together with the model (\ref{model}) and the low-level controller employed in Paparazzi for a fixed-wing aircraft.

We note that the wind has a noticeable impact on the ground speed of the aircraft. Nevertheless, as the experimental results indicate, such a wind speed does not impact the intended performance of the algorithm. In practice, we consider $\theta$ as the heading angle (given by the velocity vector), not the attitude \emph{yaw} angle. If there is no wind, both angles are the same in our setup. When we consider the heading instead of the yaw for the unicycle model (\ref{model}), the aircraft ends up compensating the lateral wind by \emph{crabbing} so that aerodynamic angle \emph{sideslip} is almost zero.\footnote{Crabbing happens when the inertial velocity makes an angle with the nose heading due to wind. \emph{Slipping} happens when the aerodynamic velocity vector makes an angle (sideslip) with the body ZX plane. Slipping is (almost) always undesirable because it degrades aerodynamic performance. Crabbing is not an issue for the aircraft.}

For tracking 3D paths (including 2D paths at constant altitude), we will employ a higher-dimensional 4D vector field. The generalized 4D velocity vector of the aircraft is defined as $\dot{\xi}=\transpose{(\dot{x}, \dot{y}, \dot{z}, \dot{w})}$, where $(\dot{x}, \dot{y})$ is the actual ground velocity of the aircraft, $\dot{z}$ is the vertical speed, and $\dot{w}$ is the velocity in the additional coordinate to be determined. 

Now we present the control algorithm design; that is, the design of $u_\theta$ and $u_z$ in \eqref{model} with the following proposition:
	\begin{prop}
	\label{prop: guidance}
		Suppose the 3D physical desired path $\phy{\setp} \subseteq \mbr[3]$ to follow is parameterized by \eqref{eq8}. Then a corresponding 4D vector field $\vf: \mbr[4] \to \mbr[4]$ can be constructed by Theorem \ref{thm2}. Assume that the vector field satisfies $\vf_1(\xi)^2 + \vf_2(\xi)^2 \ne 0$ for $\xi \in \mbr[4]$, where $\vf_i$ denotes the $i$-th entry of $\vf$.  Consider the model \eqref{model}, and let the dynamics of the additional coordinate $w$ be
		\begin{equation} \label{eq_u_w}
		\dot{w} = \frac{v \vf_4}{\sqrt{\vf_1^2 + \vf_2^2}}.
		\end{equation}
		Let the angular velocity control input $u_\theta$ and the climbing velocity input $u_z$ be
		\begin{subequations} \label{eqcontrollaw}
			\begin{align}
			u_\theta &= \underbrace{ \left(\frac{-1}{\norm{\vf^p}} \transpose{\hat{\vf^p}} E J(\vf^p) \dot{\xi}  \right)}_{\defeq \dot{\theta}_d} - k_\theta \transpose{\hat{h}} E \hat{\vf^p}, \label{eq_u_theta} \\
			u_z &= \frac{v \vf_3}{\sqrt{\vf_1^2 + \vf_2^2}}, \label{eq_u_z}
			\end{align}
		\end{subequations}
		where $k_\theta>0$ is a gain constant, $h =\transpose{(\cos \theta, \sin \theta)}$, $E=\left[\begin{smallmatrix}0 & -1 \\ 1 & 0\end{smallmatrix}\right]$, $\vf^p = \transpose{(\hat{\vf}_1, \hat{\vf}_2)}$ and $J(\vf^p)$ is the Jacobian matrix of $\vf^p$ with respect to the generalized position $\xi=(x,y,z,w)$ and $\dot{\xi}=\transpose{(\dot{x}, \dot{y}, \dot{z}, \dot{w})}$ is the generalized velocity. Let the angle difference directed from $\hat{\vf^p}$ to $\hat{h}$ be denoted by $\beta \in (-\pi, \pi]$. If the initial angle difference satisfies $\beta(0) \in (-\pi, \pi)$, then it will vanish asymptotically (i.e., $\beta(t) \to 0$). Furthermore, the actual robot trajectory $(x(t), y(t), z(t))$ will converge to the physical desired path $\phy{\setp}$ asymptotically as $t \to \infty$.
	\end{prop}
	\begin{proof}
		Let
		\[
			\vf' \defeq \frac{1}{\sqrt{\vf_1^2 + \vf_2^2}} \vf
		\]
		be the scaled 4D vector field. We aim to let the generalized robot velocity $\dot{\xi} = \transpose{(\dot{x}, \dot{y}, \dot{z}, \dot{w})}$ eventually align with and point towards the same direction as the scaled vector field. Specifically, let the orientation error be defined by
		\[
					e_{ori}(t)=\dot{\xi} - v \vf' \stackrel{\eqref{eq_u_w}, \eqref{eq_u_z}}{=} v \matr{\cos\theta -  \vf'_1 \\ \sin\theta - \vf'_2\\ 0 \\ 0} = \matr{h - g \\ \bm{0}} \in \mbr[4],
		\]
		where $h =\transpose{(\cos \theta, \sin \theta)}$ and $g=\transpose{(\vf'_1, \vf'_2)}$. It is obvious that $e_{ori} \to 0$ if and only if $h-g \to 0$. Therefore, it suffices to show that the orientation of $h$ asymptotically aligns with that of $g$. Note that 
		\[
			\hat{\vf^p}= \frac{1}{ \sqrt{\hat{\vf}_1^2 + \hat{\vf}_2^2}} \matr{\hat{\vf}_1 \\ \hat{\vf}_2}= \frac{1}{\sqrt{\vf_1^2 + \vf_2^2}} \matr{\vf_1 \\ \vf_2} = g
		\]
		and $\hat{h}=h$. Therefore, we can define a new orientation error as $e_{ori}' \defeq \hat{h} - \hat{\vf^p} \in \mbr[2]$. Choose the Lyapunov function candidate $V = 1/ 2 \, \transpose{e'}_{ori} e'_{ori}$ and its time derivative is
		\begin{equation} \label{eqvdot}
		\begin{split}
		\dot{V} = \transpose{\dot{e}'}_{ori} e'_{ori} 
		&= \transpose{( \dot{\theta} E \hat{h}  - \dot{\theta}_d E \hat{\vf^p})} (\hat{h} - \hat{\vf^p})  \\
		&= (\dot{\theta} - \dot{\theta}_d) \transpose{\hat{h}} E \hat{\vf^p} \\
		&\overset{\eqref{eq_u_theta}}{=} -k_\theta (\transpose{\hat{h}} E \hat{\vf^p})^2 \le 0.
		\end{split}
		\end{equation}
		The second equation makes use of the identities: $\dot{\hat{h}} = \dot{\theta} E \hat{h}$ and $\dot{\hat{\vf^p}} = \dot{\theta}_d E \hat{\vf^p}$, where $\dot{\theta}_d$ is defined in \eqref{eq_u_theta}. The third equation is derived by exploiting the facts that $\transpose{E}=-E$ and $\transpose{a} E a = 0$ for any vector $a \in \mbr[2]$. Note that $\dot{V}=0$ if and only if the angle difference between $\hat{h}$ and $\hat{\vf^p}$ is $\beta=0$ or $\beta=\pi$. Since it is assumed that the initial angle difference $\beta(t=0) \ne \pi$, it follows that $\dot{V}(t=0)<0$, and thus there exists a sufficiently small $\epsilon>0$ such that $V(t=\epsilon)<V(t=0)$. It can be shown by contradiction that $|\beta(t)|$ is monotonically decreasing with respect to time $t$\footnote{Suppose there exist $0<t_1<t_2$ such that $|\beta(t_1)|<|\beta(t_2)|$. It can be calculated that $V(t)=1-\cos\beta(t)$, and thus $V(t_1)<V(t_2)$, contradicting the decreasing property of $\dot{V}$. Thus $|\beta(t)|$ is indeed monotonically decreasing.}. By \eqref{eqvdot}, one observes that $|\beta(t)|$ and $V(t)$ tends to $0$, implying that the generalized velocity $\dot{\xi}$ will converge asymptotically to the scaled vector field $v \vf'$. Note that the integral curves of the state-dependent positive scaled vector field $\vf'$ has the same convergence results as those for the original vector field $\vf$ \cite[Proposition 1.14]{chicone2006ordinary}. Therefore, the generalized trajectory $(x(t), y(t), z(t), w(t))$ will converge to the higher-dimensional desired path $\hgh{\setp}$ in \eqref{eq_high_p}. From Theorem \ref{thm2}, the actual robot trajectory (i.e., the projected transformed trajectory)  $(x(t), y(t), z(t))$ will converge to the physical desired path $\phy{\setp}$ asymptotically as $t \to \infty$.
	\end{proof}

\begin{figure*}[!t]
	\small
	\begin{equation}
	J(\vf) = L\matr{
		-k_1L             &    0                 & 0                & - f_1''(\beta w)L^2\beta^2 + k_1\beta L f_1'(\beta w) \\
		0             &    -k_2L                 & 0                & - f_2''(\beta w)L^2\beta^2 + k_2\beta L f_2'(\beta w) \\
		0                 &    0                 & -k_3L            &  - f_3''(\beta w)L^2\beta^2 + k_3\beta L f_3'(\beta w) \\
		k_1\beta L f_1'(\beta w)  &   k_2\beta L f_2'(\beta w)  &  k_3\beta L f_3'(\beta  w) & \beta^2 \sum_{i=1}^{3} \left[ k_i \phi_i f_i''(\beta w) - k_i L f_i'^2(\beta w) \right] }
	\label{eq: Jx}
	\end{equation}
	\hrulefill
	\vspace*{4pt}
	\normalsize
\end{figure*}

We set our aircraft to fly at constant airspeed (around $12m/s$) while flying at constant altitude; therefore, we have a bounded speed $v$ (estimated onboard with an Inertial Navigation System) when we account for the wind. For tracking 3D paths, the aircraft will nose down or change the propeller's r.p.m.; nevertheless, the airspeed is also bounded between $9 m/s$ and $16 m/s$. Note that both ground and airspeed are not control/guiding signals; therefore, we do not face any saturation problems regarding these variables.

\subsection{Accommodating the guidance to the aircraft's dynamics}
\label{sec: prac}

An arbitrary function $\phi_i(\cdot)$ in \eqref{eqsurface1}, which depends on a specific parametrization $f_i(\cdot)$, may result in a highly sensitive coordinate $w$. This can lead to considerable vibrations of the guidance signals, due to noisy sensor readings or disturbances of the position, that cannot be tracked effectively by the aircraft.

We propose two approaches, which can be combined to mitigate this practical effect. The first one is to re-parameterize the equations for the 3D desired path $\phy{\setp}$; this does not affect the convergence result. Suppose $\phy{\setp}$ is re-parameterized by
	\[
	x = f_1 \big(g(w) \big), \quad y = f_2 \big(g(w) \big), \quad z = f_3 \big(g(w) \big),
	\]
	where $g: \mbr[] \to \mbr[]$ is a smooth bijection with nonzero derivative (i.e., $\derivative{g}{w}(w) \ne 0$ for all $w \in \mbr[]$). A simple example of $g$ is $g(w)=\beta w$, where $\beta$ is a positive constant. This is adopted for the experiments. Let $\phi_1$, $\phi_2$, and $\phi_3$ be chosen as in \eqref{eqsurface1}, then the first term of the higher-dimensional vector field becomes (for simplicity, the arguments are omitted)
	\begin{equation} \label{eq18}
	\times (\nabla \phi_1, \nabla \phi_2, \nabla \phi_3)  = - \matr{\derivative{f_1}{g} \cdot \derivative{g}{w} \\ \derivative{f_2}{g} \cdot \derivative{g}{w} \\ \derivative{f_3}{g} \cdot \derivative{g}{w} \\ 1}.
	\end{equation}
	To reduce the effect of the ``virtual speed'' from the fourth coordinate of \eqref{eq18}, the ``gain'' $\derivative{g}{w}$ can be chosen large such that $(\derivative{f_1}{g} \cdot \derivative{g}{w})^2 + (\derivative{f_2}{g} \cdot \derivative{g}{w})^2 + (\derivative{f_3}{g} \cdot \derivative{g}{w})^2 \gg 1$, which implies that $\norm{\nabla \phi_1 \times \nabla \phi_2 \times \nabla \phi_3} \approx \left|\derivative{g}{w}\right| \sqrt{(\derivative{f_1}{g})^2 + (\derivative{f_2}{g})^2 + (\derivative{f_3}{g})^2}$. However, from the analytic expression of the vector field
	\[	
	\vf = \matr{-\derivative{g}{w} \cdot \derivative{f_1}{g} - k_1 \phi_1 \\ -\derivative{g}{w} \cdot \derivative{f_2}{g}  - k_2 \phi_2 \\ -\derivative{g}{w} \cdot \derivative{f_3}{g} - k_3 \phi_1 \\ -1 + \derivative{g}{w} \left( k_1 \phi_1 \derivative{f_1}{g} + k_2 \phi_2 \derivative{f_2}{g} + k_3 \phi_3 \derivative{f_3}{g} \right)},
	\]
one observes that, when $\norm{(\phi_1, \phi_2, \phi_3)}$ is large, (i.e., the aircraft is far from the desired path), the additional coordinate of the vector has also been enlarged approximately by a factor of $\derivative{g}{w}$. Thus, the ``gain'' $\lvert \derivative{g}{w} \rvert$ should not be chosen too large. 
	
The second approach is to scale down the functions $\phi_i$. That is, the equations \eqref{eqsurface1} are changed to
	\begin{equation} \label{eq17_2}
		\phi_i(x,y,z,w) = L( x_i - f_i(w)),   i=1,2,3, \nonumber
	\end{equation}
where $L \in (0,1)$. The corresponding 3D vector field is thus changed to
	\[
	\tilde{\vf} = L \matr{ -L^2 \derivative{f_1}{w} - k_1 \phi_1 \\ -L^2 \derivative{f_2}{w} - k_2 \phi_2 \\ -L^2 \derivative{f_3}{w} - k_3 \phi_3 \\ -L^2 + k_1 \phi_1 \derivative{f_1}{w} + k_2 \phi_2 \derivative{f_1}{w} +  k_3 \phi_3 \derivative{f_1}{w} }.
	\]
The new guiding vector field is scaled down; thus, it helps to lower the sensitivity of the additional coordinate $w$.

\subsection{The 2D trefoil curve}
We start with following a 2D self-intersected desired path, the trefoil curve, at a constant altitude $z_o = 50m$ over the ground level. The parametric equations of the trefoil curve are given by
\begin{align}
	f_1(w) &= \cos(\beta w \,\omega_1)(a\cos(\beta w \,\omega_2) + b) \nonumber \\
	f_2(w) &= \sin(\beta w \,\omega_1)(a\cos(\beta w \,\omega_2) + b) \nonumber \\
	f_3(w) &= 0, \nonumber
\end{align}
where we have set $\beta = \derivative{g}{w} = 0.45$ (the ``gain" introduced in Section \ref{sec: prac}), $\omega_1 = 0.02$, $\omega_2 = 0.03$, $a = 80$, and $b=160$. In order to fit into the available flying space, these parametric equations have been rotated by $\alpha$ and shifted adequately by $(x_o,y_o)$ in the autopilot, i.e., 
\begin{align}
	f_1^*(w) &= \cos(\alpha)f_1(w) -\sin(\alpha)f_2(w) + x_o \nonumber \\
	f_2^*(w) &= \sin(\alpha)f_1(w) +\cos(\alpha)f_2(w) + y_o \nonumber \\
	f_3^*(w) &= f_3(w) + z_o. \nonumber
\end{align}
Note that the same affine transformation must be done for both $f_i'$ and $f_i''$ (needed for the Jacobian of $\vf$ as we will see shortly). In particular, for the presented flight, we have set $x_o=79$, $y_o= -68.10$ and $z_o=50$ meters and $\alpha=0$. We set the scaling factor $L = 0.1$ for the construction of $\tilde\phi_i$ as in Section \ref{sec: prac}, and we have chosen the gains $k_1 = k_2 = k_3 = 0.002$. We finally set $k_\theta = 1$ for the control/guidance signal $u_\theta$ in Proposition \ref{prop: guidance}.

\begin{figure*}
\centering
\includegraphics[width=2\columnwidth]{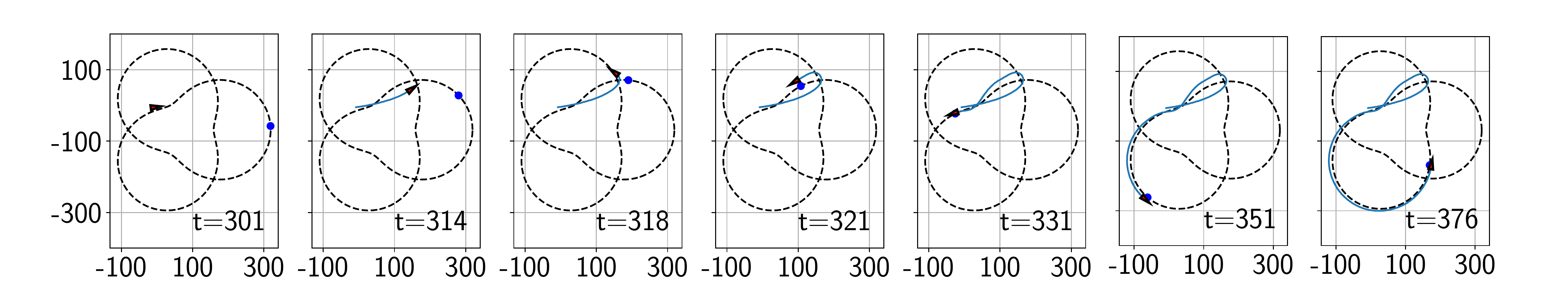} \\
\includegraphics[width=1.1\columnwidth]{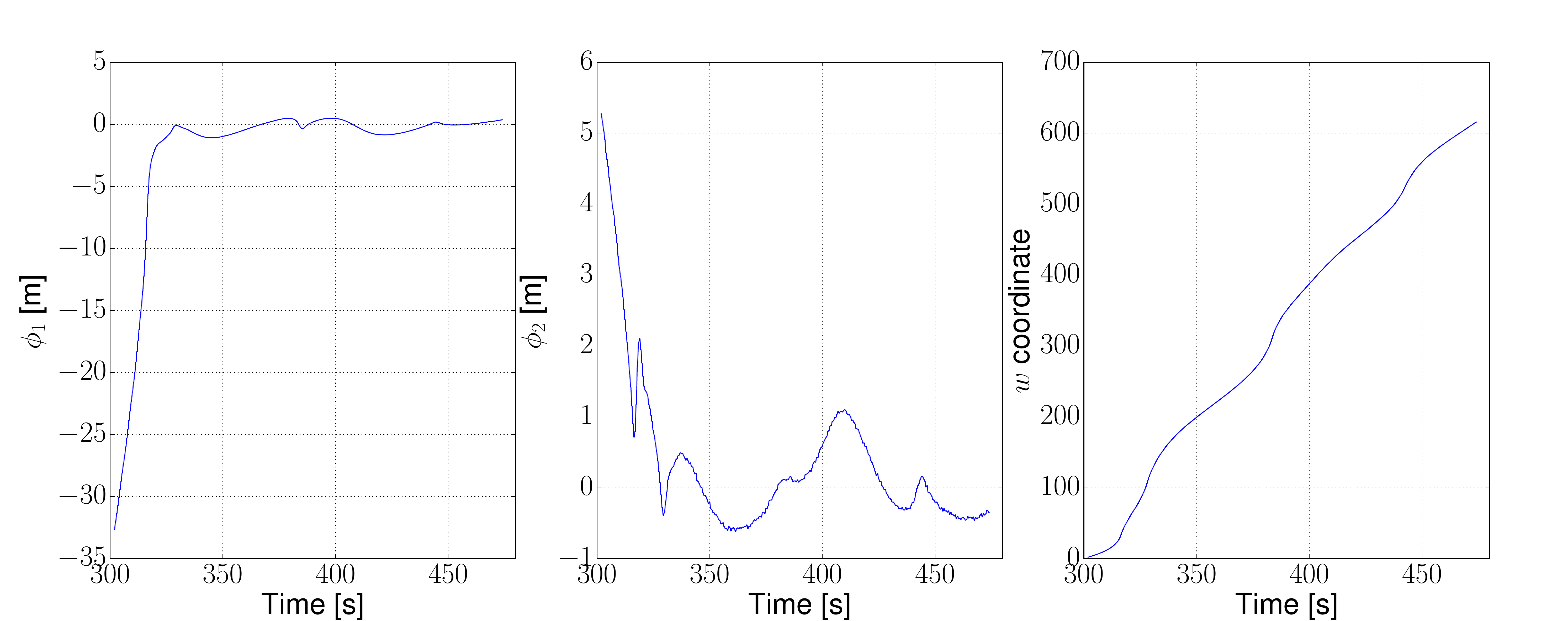}
\includegraphics[width=0.9\columnwidth]{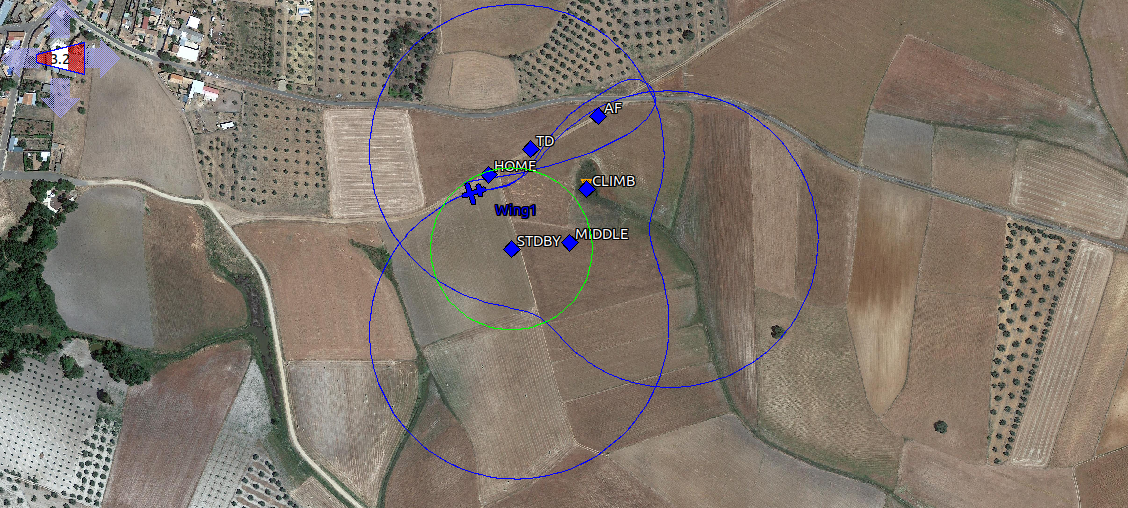}
	\caption{Flight results of the trefoil path tracking at a constant altitude. On the low-right side, we show a screenshot of the ground control station, where the green circle denotes the \emph{stand-by} trajectory just before starting the tracking of the trefoil. On top, we show the first minute of the flight once the tracking of the trefoil started. All the tracking is done at the constant altitude of $50$ meters, and the evolution of the coordinate $z$ is thus omitted. We can see how the blue dot, i.e., $(f_1(w),f_2(w))$ goes and \emph{waits} for the aircraft at time $t=321$. Afterward, the vehicle converges to the desired path as the first two low-left plots indicate with $\phi_1$ and $\phi_2$ staying around $0$ meters. The third plot shows the evolution of the virtual coordinate $w$. We can see how $w$ does not follow a constant growth, but it varies since it is in closed-loop with the position of the aircraft to facilitate the convergence to the path. Coming back to the ground control station, we can see in blue color the described 2D trajectory of the aircraft, showing one period of the trefoil.}
\label{fig: trefoil}
\end{figure*}

Note that for computing all the control signals \eqref{eqcontrollaw}, we need $f_i(w)$ and their derivatives $f_i'(w)$ and $f_i''(w)$ with respect to $w$. For the sake of completeness, we provide the Jacobian $J(\vf^p)$ in \eqref{eq_u_theta} which is given by
\begin{equation}
J(\vf^p)=F J(\hat{\vf})= F(I-\hat{\vf} \transpose{\hat{\vf}}) J(\vf) / \norm{\vf},
\end{equation}
where $F = \left[\begin{smallmatrix}1 & 0 & 0 & 0 \\ 0 & 1 & 0 & 0\end{smallmatrix}\right]$, and $J(\vf)$ is shown in (\ref{eq: Jx}). We show the flight results for the trefoil curve in Figure \ref{fig: trefoil}.

\subsection{The 3D Lissajous curve}
\begin{figure*}
\centering
\includegraphics[width=2\columnwidth]{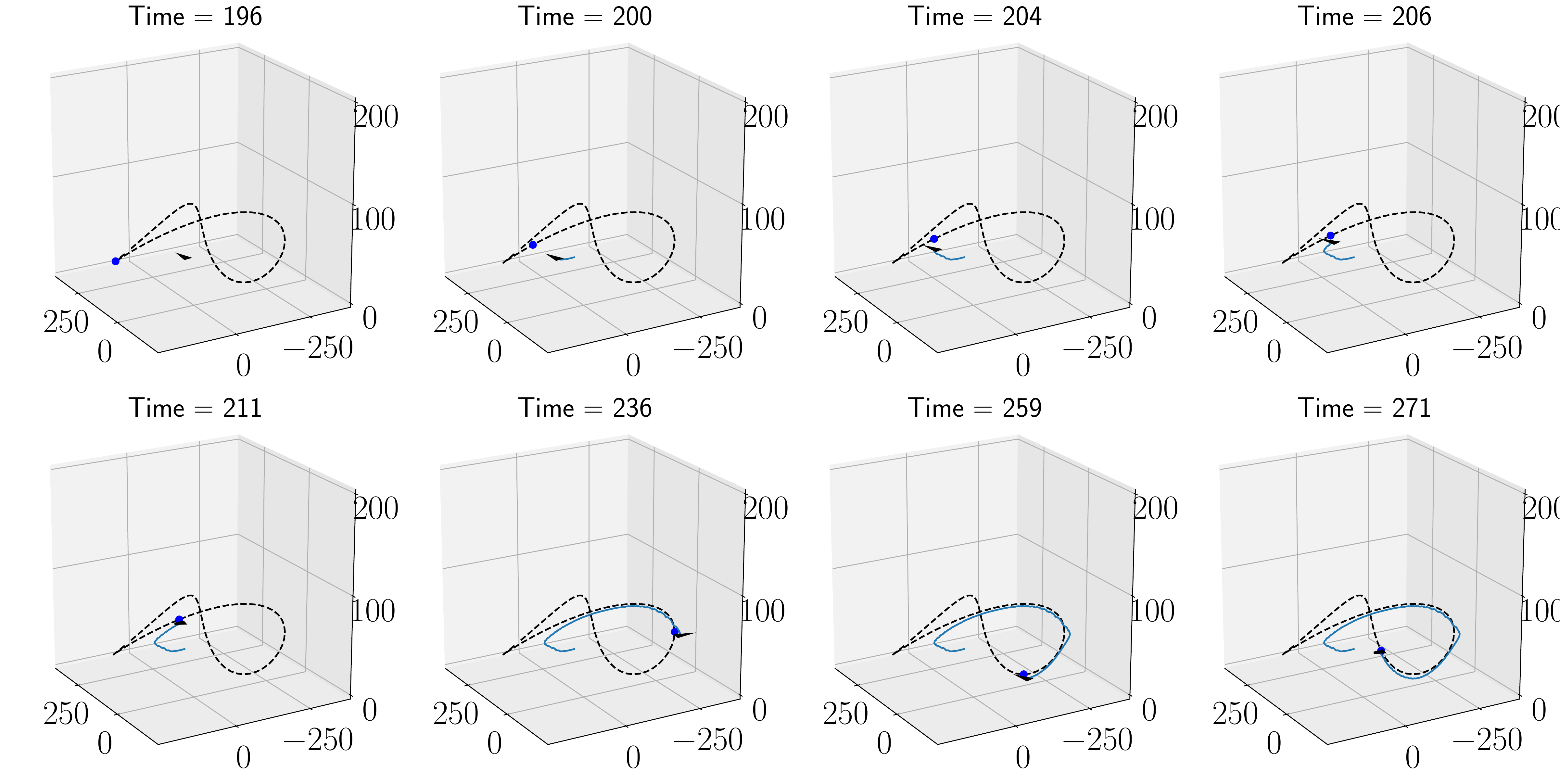} \\
\includegraphics[width=1.1\columnwidth]{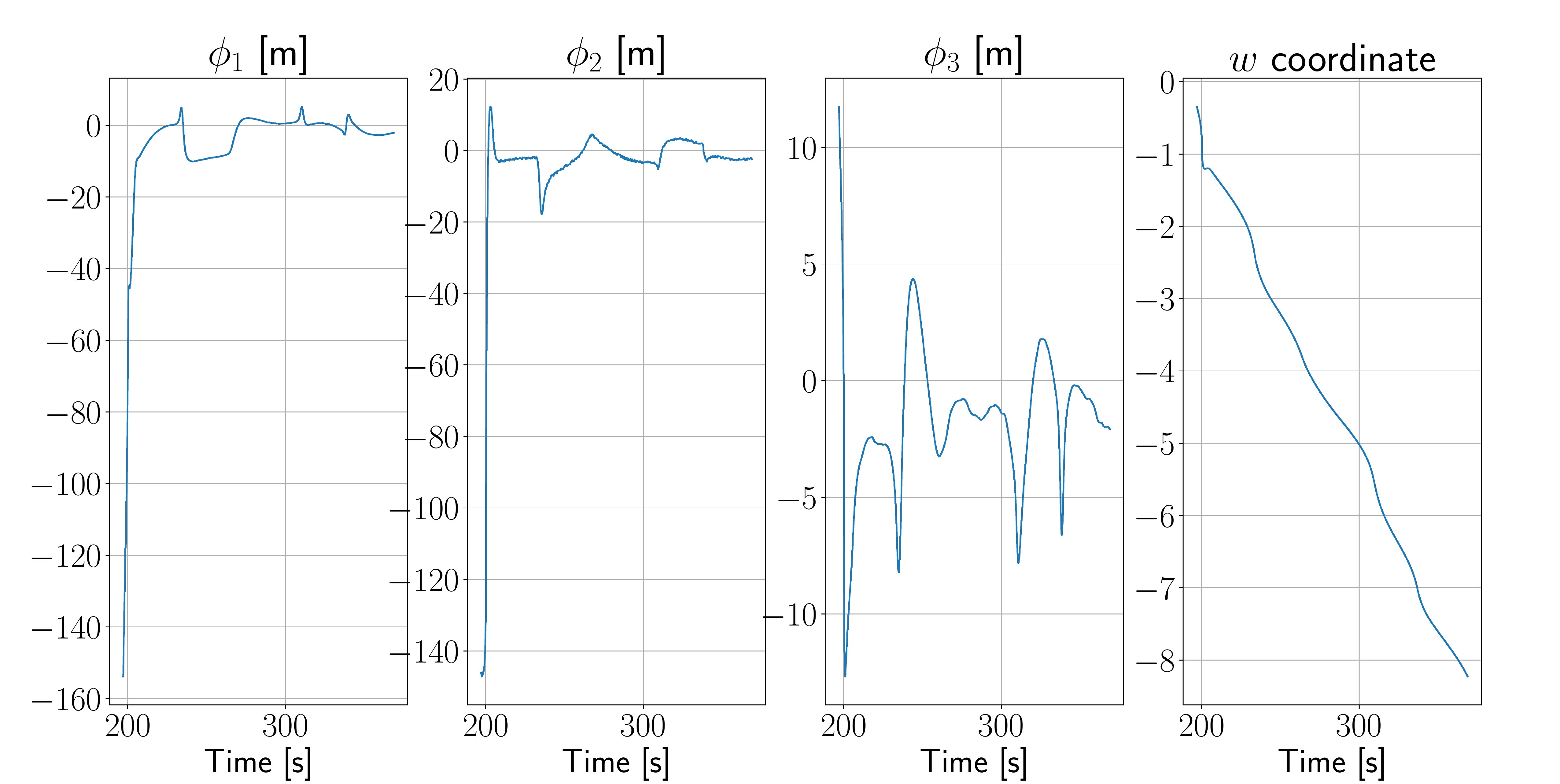}
\includegraphics[width=0.9\columnwidth]{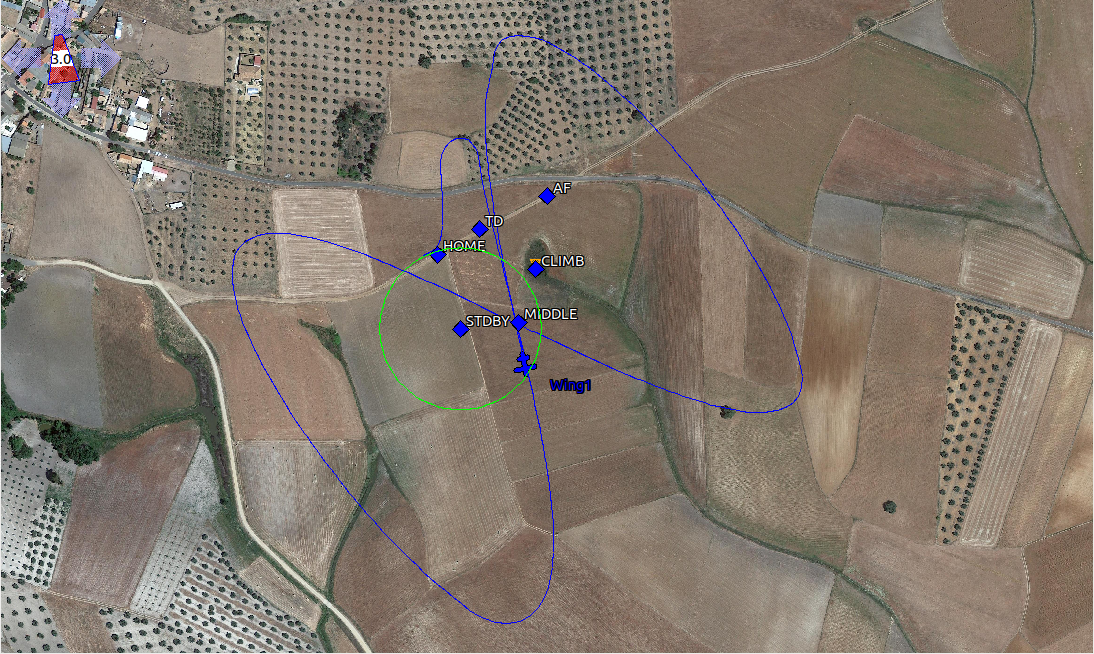}
	\caption{Flight results of the 3D Lissajous path tracking. On the low-right side, we show a screenshot of the ground control station, where the green circle denotes the \emph{stand-by} trajectory just before starting the tracking of the Lissajous figure. In the 3D plots on top, we show the first seconds of the flight once the tracking of the Lissajous figure started. We can see how the blue dot, i.e., $(f_1(w),f_2(w),f_3(w))$ travels quickly and \emph{waits} for the aircraft at time $t=204$. This quick traveling of $w$ can be seen at the beginning of the fourth plot in the bottom row. Afterward, the vehicle stays on the desired path as the three bottom-left plots indicate with $\phi_1, \phi_2$, and $\phi_3$ staying around $0$ meters. The aircraft has been trimmed to fly at a constant altitude. However, this path requires the vehicle to track a sinusoidal ascending/descending trajectory, and any disturbance (e.g., changing wind) makes the aircraft sensitive to track a climbing/descending speed different from zero accurately. In addition, the Lissajous curve demands aggressive turnings a bit beyond the capabilities of the aircraft, in particular, when the aircraft is descending and achieving maximum speed. We can see how $w$ does not follow a constant growth, but it varies since it is in closed-loop with the position of the aircraft to facilitate the convergence to the path. Coming back to the ground control station, we can see in blue color the described 2D (projected) trajectory of the aircraft, showing one period of the Lissajous figure. In particular, we can see that the aircraft passes by the \emph{middle waypoint} corresponding to the highest point of the desired track.}
\label{fig: lissa}
\end{figure*}

We consider the 3D Lissajous curve described as below:
\begin{align}
	f_1(w) &= c_x\cos(\beta w \, \omega_x + d_x) \nonumber \\
	f_2(w) &= c_y\cos(\beta w \, \omega_y + d_y) \nonumber \\
	f_3(w) &= c_z\cos(\beta w \, \omega_z + d_z), \nonumber
\end{align}
where we have set $\beta = \derivative{g}{w} = 0.01$, $\omega_x = 1$, $\omega_y = \omega_z = 2$, $c_x = c_y = 225$, $c_z = -20$, $d_x = d_z = 0$, and $d_y = \pi/2$. This selection of parameters gives us an \emph{eight-shaped} desired path that is bent along the vertical axis. As with the trefoil curve, we have added an affine transformation of $f_i(w), f_i'(w)$ and $f_i''(w)$  in the autopilot to fit the Lissajous curve into the available flying space. In particular, we have set $x_o=79, y_o=-68.10, z_o=50, \alpha= 0.66$). Finally, for the construction of $\tilde\phi_i$, we have chosen $L = 0.1$, $k_1 = k_2 = 0.002$ and $k_3 = 0.0025$. We finally set $k_\theta = 1$ for the control/guiding signal $u_\theta$ in Proposition \ref{prop: guidance}. We show the flight results in in Figure \ref{fig: lissa}.

\section{Discussion \\ Path following or trajectory tracking?} \label{sec_disc}
In this section, we show that our proposed higher-dimensional VF-PF algorithm is an extension that combines elements from both conventional VF-PF algorithms (e.g., \cite{kapitanyuk2017guiding}, see Remark \ref{remark_variants}) and trajectory tracking algorithms (e.g., \cite[p. 506]{siciliano2010robotics}). While our generated guiding vector field is the standard output for the path-following approach, we will argue that our algorithm can also be seen as a fair extension of a trajectory tracking approach. Therefore, our algorithm, to some extent, combines and extends elements from both approaches. For ease of explanation and without loss of generality, we restrict our focus to a physical planar desired path in $\mbr[2]$; that is, $\phy{\setp} \subseteq \mbr[2]$.
	
Compared to trajectory tracking algorithms, a similarity exists in the sense that the additional coordinate $w$ in the proposed VF-PF algorithms acts like the time variable in trajectory tracking algorithms. However, our approach is an extension in the sense that the time-like variable is in fact state-dependent. In trajectory tracking algorithms, a desired trajectory $\big( x_d(t), y_d(t) \big)$ is given. Then, at any time instant $t$, the algorithm aims to decrease the distance to the \emph{desired trajectory point} $\big( x_d(t), y_d(t) \big)$, which moves as time $t$ advances. Note that the dynamics of the \emph{desired trajectory point} $\big( x_d(t), y_d(t) \big)$ is \emph{open-loop} in the sense that it does not depend on the current states of the robot, but only depends on time $t$. From \eqref{eqsurface1}, if we let $\phi_i=0$, $i=1,2$, then we may call the point $\big( f_1(w(\xi(t) ) ), f_2 (w(\xi(t) ) ) \big)$ the \emph{guiding point}, since it always stays on the desired path and may be regarded as the counterpart of the \emph{desired trajectory point} in trajectory tracking algorithms. But as we will show later, the \emph{guiding point} is essentially different from the \emph{desired trajectory point}. Note that the \emph{guiding point} $\big( f_1(w(\xi(t) ) ), f_2 (w(\xi(t) ) ) \big)$ in our VF-PF algorithm depends on the evolution of the additional coordinate $w \big( \xi(t) \big)$, of which the dynamics is state-dependent as shown in \eqref{eq_u_w}. This might be roughly regarded as a \emph{closed-loop} version of the \emph{desired trajectory point}. An intuitive consequence of this difference is that the \emph{desired trajectory point} $\big( x_d(t), y_d(t) \big)$ in trajectory tracking algorithms always moves \emph{unidirectionally} along the desired trajectory as $t$ monotonically increases, while the \emph{guiding point} can move \emph{bidirectionally} along the desired path, subject to the current state (i.e., the robot position). In fact, when the initial position of the \emph{guiding point} $\big( f_1(w(\xi(0) ) ), f_2 (w(\xi(0) ) ) \big)$ is far from the initial position of the robot, the \emph{guiding point} ``proactively'' moves towards the robot along the desired path to accelerate the path-following process. This feature, along with better robustness against perturbation in some cases, are experimentally studied in our previous work \cite[Section VII]{yao2020mobile2}. To illustrate this closed-loop feature more intuitively, after the robot has successfully followed the desired path, we manually move the robot far away from the desired path and keep it stationary (to mimic the situation of erroneous localization and operation failure of the robot). As is clear in the supplementary video\footnote{\url{https://www.youtube.com/watch?v=jxWPsm0g-Ro}.}, although the robot is kept stationary, the guiding point $\big( f_1(w(\xi(t) ) ), f_2 (w(\xi(t) ) ) \big)$ can still move in the reverse direction to approach the robot along the desired path such that the path-following error decreases, and the guiding point eventually stops at some place on the desired path. After that, the guiding point does not move until the robot is released to move again. 
	
\begin{figure}[tb]
		\centering
		\includegraphics[width=1\linewidth]{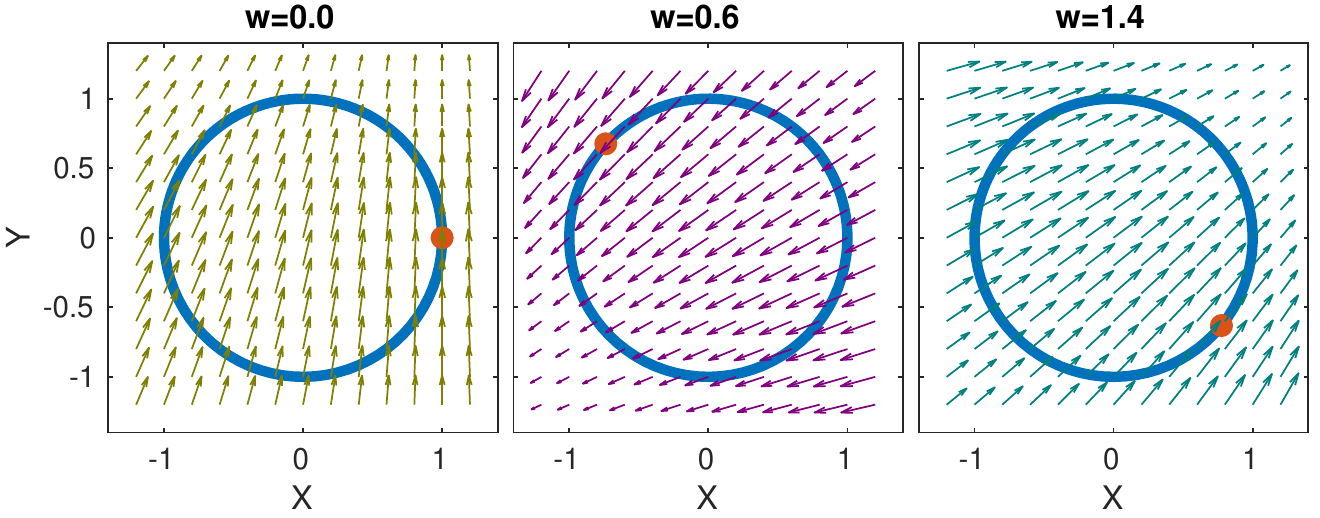}
		\caption{The projected 2D vector field corresponding to $w=0, 0.6, 1.4$ respectively. The solid line is the projected 2D desired path. The solid dots represent the guiding point $\big( cos(4w), sin(4w) \big)$.}
		\label{fig:discuss2}
	\end{figure}

In existing VF-PF algorithms, a two-dimensional vector field on $\mbr[2]$ is created for guiding the robot movement (see Remark \ref{remark_variants}). However, as we aim to create a higher-dimensional (i.e., three-dimensional) vector field, our approach can be roughly regarded as utilizing an infinite number of layers of projected two-dimensional vector fields, and thus might be seen as a \emph{dynamic} two-dimensional vector field. The dynamic property is due to the dynamics of the additional coordinate $w$. For example, consider a circular desired path parameterized by 
	\[
	x=f_1(w)=\cos(4w) \quad y=f_2(w)=\sin(4w),
	\]
where $w \in \mbr[]$ is the parameter. In conventional VF-PF algorithms, a 2D vector field can be created, as shown in Fig. \ref{fig:circlevf}, but there exists a singular point at the center of the circle. Nevertheless, using our approach, we can generate a singularity-free 3D vector field, as illustrated in Fig. \ref{fig:discuss1}. For clarity of visualization, we plot the 3D vectors, which originate from three planes where the $w$ values are $0, 0.6$, and $1.4$, respectively. For each value of the additional coordinate $w$, we can obtain a projected 2D vector field, as shown in Fig. \ref{fig:discuss2}. Therefore, we can observe that these 2D vector fields change dynamically as $w$ varies. As a result of the dynamics of $w$, the \emph{guiding point} $\big( f_1(w(\xi(t))), f_2(w(\xi(t))) \big)$ moves along the 2D desired path (not necessarily uni-directionally). Again, we note that this point is \textbf{not} the same as the \emph{desired trajectory point} in trajectory tracking algorithms since the integral curves of the 2D vector field do not converge to this point, as can be seen graphically from Fig. \ref{fig:discuss2} or analytically from the expression of the vector field in \eqref{eqvf2}: the second term leads to convergence to the \emph{guiding point}, while the first term ``deviates'' this convergence, since it controls the propagation along the \emph{higher-dimensional} desired path.
	
In many existing VF-PF algorithms, the desired path is usually not parameterized but is described by the intersection of (hyper)surfaces, while the latter case might be restrictive in describing more complicated desired paths. However, our approach enables the possibility to use a parameterized desired path directly in the design of a higher-dimensional vector field. Our approach thus extends the flexibility of conventional VF-PF algorithms. The desired path can now be described by either the intersection of (hyper)surfaces or parameterized functions. In the latter case, the parametric equations can be easily converted using \eqref{eq8}, \eqref{eqsurface1} and \eqref{eq_high_p} and leads to a higher-dimensional desired path and singularity-free guiding vector field. Theoretically, the parametrization is not instrumental, since it is only utilized to derive the expressions of functions $\phi_i$, of which the zero-level sets are interpreted as (hyper)surfaces. The subsequent derivation of the vector field is based on $\phi_i$, independent of the specific parametrization of the desired path.

	\begin{figure}[tb]
		\centering
		\includegraphics[width=0.8\linewidth]{fig/discuss1-eps-converted-to.pdf}
		\caption{Three layers of the 3D vector field corresponding to a circle. The solid line is the 2D desired path while the dashed line is the corresponding 3D (unbounded) desired path. Three layers of the 3D vector field evaluated at $w=0, 0.6, 1.4$ respectively are illustrated. }
		\label{fig:discuss1}
	\end{figure}
	
	\section{Conclusion and Future work} \label{sec_conclu}
	In this paper, we first show that the integral curves of a time-invariant continuously differential vector field as in \eqref{gvf} cannot guarantee global converge to desired paths which are simple closed (i.e., homeomorphic to the unit circle) or self-intersected. Motivated by this general topological result,  we propose a novel approach to create unbounded desired paths from simple closed or self-intersected ones, and construct a singularity-free higher-dimensional guiding vector field. One of the advantages of this approach is that global convergence to the desired paths, which can be even self-intersected, is now rigorously guaranteed. This is achieved by the introduction of a transformation operator and the extended dynamics. Another advantage is that, given a parameterized desired path, we can easily describe the (hyper)surfaces as the zero-level set of some implicit functions, and then the proposed  vector field on a higher-dimensional space can be directly constructed. This increases the applicability of conventional VF-PF algorithms. In addition, we highlight five features of our approach, with rigorous theoretic guarantees. We also show that our approach is a combined extension of both conventional VF-PF algorithms and trajectory tracking algorithms. Finally, we conduct outdoor experiments with a fixed-wing aircraft under wind perturbation to validate the theoretical results and demonstrate the practical effectiveness for complex robotic systems. 
	
	Due to the additional coordinates of the vector field, it is difficult for a robot to follow the desired path at a constant speed. This may be solved by replacing the normalization of the vector field $\hat{\vf}$ by the ``partial normalization''; i.e., $\tilde{\vf} \defeq \vf / \norm{\vf_p}$, where $\vf_p$ is the vector composed of only the first $(n-1)$ components of $\vf$ (the $n$-th component is the additional coordinate). It is our future work to use the singularity-free guiding vector field to solve collision avoidance problems; preliminary results are reported in \cite{yao2019integrated}. We are also interested in determining the parameterization of the desired path to achieve the optimal path-following accuracy.
	
	\section*{Acknowledgments}
	Weijia Yao and Bohuan Lin are funded by the China Scholarship Council. The work of H{\' e}ctor Garcia de Marina is supported by the grant Atraccion de Talento 2019-T2/TIC-13503 from the Government of the Autonomous Community of Madrid. The work of Hector is also partially supported by the Spanish Ministry of Science and Innovation under research Grant RTI2018-098962-B-C21. The work of Cao was supported in part by the European Research Council (ERC-CoG-771687) and the Netherlands Organization for Scientific Research (NWO-vidi-14134). We thank Murat Bronz and Gautier Hattenberger from the \emph{Paparazzi} project for their assistance and feedback. We also thank Yangguang Yu and Kyle Miller for the technical discussions. 
	
	\appendices
	\section{Proofs and Supplementary Material for Section \ref{sec_vf}} \label{app0}
	\subsection{Proof of Lemma \ref{lemma_mpc}} \label{appA1}
	\begin{proof}
		First, it is easy to see that for any point $\xi \in \setp \cup \setc$, we have $\xi \in \setm$, thus $\setp \cup \setc \subseteq \setm$. Second, for any point $\xi' \in \setm$, it follows that $N(\xi') K e(\xi') = \sum_{i=1}^{n-1} k_i \phi_i(\xi') \nabla \phi_i(\xi') = 0$. If $e(\xi')=0$, then $\xi' \in \setp$. If $e(\xi') \ne 0$, then the former equation implies that $\nabla \phi_i(\xi'), \; i=1,\dots,n-1$, are linearly dependent (recalling that $k_i>0$); hence the first term of the vector field becomes zero (i.e., $\nabla_{\times} \phi(\xi)=0$). Since $\xi' \in \setm$, the second term of the vector field is also zero, thus the vector field $\vf(\xi')=0$ and $\xi' \in \setc$. The reasoning shows that $\setm \subseteq \setp \cup \setc$. Combining $\setp \cup \setc \subseteq \setm$ and $\setm \subseteq \setp \cup \setc$, it is indeed true that $\setm = \setp \cup \setc$. 
	\end{proof}
	
	\subsection{Proof of Lemma \ref{lemmanifold}} \label{appA2}
	\begin{proof}
		By the definition of the desired path $\setp$ in \eqref{path2}, it is obvious that the desired path is the inverse image of $e(\bm{0})$; that is, $\mathcal{P}=e^{-1}(\bm{0})$. Under Assumption \ref{assump1}, for any $\xi \in \mathcal{P}$, the Jacobian matrix $\mathrm{d}\, e / \mathrm{d}\, \xi =  \transpose{N}(\xi)$ is of full row rank. Thus the result follows directly from \cite[Corrollary 5.14]{lee2015introduction}.
	\end{proof}
	
	\subsection{Mathematical Formulation of Assumption \ref{assump2}} \label{appA3}
	To keep the paper self-contained, we restate Assumption \ref{assump2} using precise mathematical descriptions. The first part is: for any given constant $\kappa > 0$, it follows that $\inf\{ || e(\xi)||: \dist(\xi, \mathcal{P}) \ge \kappa\} > 0$. Under this assumption, it can be proved that as the norm of the path-following error $\norm{e(\cdot)}$ vanishes, the distance to the desired path $\dist(\cdot, \setp)$ also vanishes; that is, $\lim\limits_{t \rightarrow \infty} ||e \big( p(t) \big)||=0 \implies \lim\limits_{t \rightarrow \infty}  \dist(p(t), \mathcal{P})=0$ \cite{yao2020auto,yao2021dichotomy}. The second part of the assumption is: for any given constant $\kappa > 0$, it holds that $	\inf \left\{ \norm{ N(\xi) K e(\xi)}:  \dist(\xi, \mathcal{M}) \ge \kappa \right\} > 0$. Similarly, this implies $\lim\limits_{t \rightarrow \infty} ||N \big( p(t) \big) K e \big( p(t) \big)||=0 \implies \lim\limits_{t \rightarrow \infty}  \dist(p(t), \mathcal{M})=0$ \cite{yao2020auto,yao2021dichotomy}.

	\section{Proofs for Section \ref{sec_impos}} \label{app1}
	\subsection{Proof of Proposition \ref{prop1}} \label{appB1}
	\begin{proof}
		Since $c \in \mathcal{P}$ is a crossing point, we have $e(c)=0$, and thus the vector field at the crossing point is simplified to $\vf(c) = \nabla_{\times} \phi(c)$ in view of \eqref{eqvf2}. Next we show that the gradients at the crossing point $\nabla \phi_i(c)$ are linearly dependent, and hence $\vf(c)=0$. Suppose, on the contrary, the gradients are not linearly dependent. Then we can use the implicit function theorem \cite{giaquinta2010mathematical} to conclude that there is a unique curve in a neighborhood of $c$ satisfying $e(\xi)=0$, where $\xi \in \mbr[n]$. But this contradicts the fact that $\setp$ is self-intersected. Therefore, the gradients at the crossing point are indeed linearly dependent. 
	\end{proof} 
	
	\subsection{Proof of Proposition \ref{note_prop2new}} \label{appB2}
	Given an autonomous differential equation $\dot{x}(t)=f(x(t))$, where $f$ is continuously differentiable in $x$, and let $t \mapsto \Psi(t,x_0)$ be the solution to the differential equation with the initial condition $\Psi(0,x_0)=x_0$, then $\Psi$ is a flow \cite{chicone2006ordinary}. In the literature, the notation $\Psi^t(x_0)$, which is adopted in the sequel, is often used in place of $\Psi(t,x_0)$. To assist the proof of Proposition \ref{note_prop2new}, we state a more general result in the following lemma regarding any time-invariant autonomous system that admits a (locally) asymptotically stable limit cycle. Note that similar to the definition of Lyapunov stability of an equilibrium point \cite[Chapter 4]{khalil2002nonlinear}, a limit cycle $\mathcal{L}$ is (locally) asymptotically stable if for every neighborhood $\mathcal{U} \supseteq \mathcal{L}$ of the limit cycle $\mathcal{L}$, there exists a \emph{smaller} neighborhood $\mathcal{V} \subseteq \mathcal{U}$, such that every trajectory starting from $\mathcal{V}$ always stays within $\mathcal{U}$ and $\mathcal{L}$ is locally attractive. The proof of this lemma involves notions about topological and differential manifolds \cite{lee2015introduction, lee2010topologicalmanifolds}, which are briefly explained here. 	If $\set{V}$ and $\set{W}$ are vector spaces, a bijective linear map $T: \set{V} \to \set{W}$ is called an \emph{isomorphism}. If this isomorphism exists, then $\set{V}$ and $\set{W}$ are called \emph{isomorphic}, denoted by $\set{V} \cong \set{W}$. Suppose $\set{X}, \set{Y}$ are topological spaces. An \emph{open cover} of $\set{X}$ is a collection of open subsets of $\set{X}$ whose union is $\set{X}$. A \emph{diffeomorphism} $f: \set{X} \to \set{Y}$ is a smooth bijection that has a smooth inverse. If there exists a diffeomorphism between $\set{X}$ and $\set{Y}$, then $\set{X}$ and $\set{Y}$ are \emph{diffeomorphic}. Let $f,g: \set{X} \to \set{Y}$ be continuous maps. \emph{A homotopy from $f$ to $g$} is a continuous map $H: \set{X} \times [0,1] \to \set{Y}$ such that $H(x,0)=f(x)$ and $H(x,1)=g(x)$ for all $x \in \set{X}$. If there exists such a homotopy, then $f$ and $g$ are \emph{homotopic}, denoted by $f \simeq g$.	Let $\set{A} \subseteq \set{X}$. The \emph{inclusion map of $\set{A}$ in $\set{X}$} is $\iota_{\set{A}}: \set{A} \to \set{X}$ defined by $\iota_{\set{A}}(x)=x$ for $x \in \set{A}$. A continuous map $r: \set{X} \to \set{A}$ is a \emph{retraction} if the restriction of $r$ to $\set{A}$ is the identity map of $\set{A}$, or equivalently if $r \circ \iota_{\set{A}} = \identity_{\set{A}}$, where $\iota_{\set{A}}$ is the inclusion map of $\set{A}$ in $\set{X}$, and $\identity_{\set{A}}$ is the identity map of $\set{A}$. In this case, $\set{A}$ is called a \emph{retract of $\set{X}$}. The space $\set{X}$ is called \emph{contractible} if the identity map $\identity_\set{X}$ of $\set{X}$ is homotopic to  a constant map. Intuitively, this means that the whole space $\set{X}$ can continuously shrink to a point. For the definition of a tubular neighborhood and an introduction to fundamental groups and homomorphisms of fundamental groups induced by continuous maps, see \cite[pp. 137-139]{lee2015introduction} and \cite[Chapter 7]{lee2010topologicalmanifolds} respectively.
	
	\begin{lemma}[Asymptotically stable limit cycles are not GAS] \label{lemma3_new}
		Consider an autonomous differential equation $\dot{x}(t) = f(x(t))$, where $f: \mbr[n] \to \mbr[n]$ is continuously differentiable in $x$. Suppose there is a (locally) asymptotically stable limit cycle $\mathcal{L} \subseteq \mbr[n]$, then global convergence of trajectories to the limit cycle is not possible; namely, the domain of attraction of the limit cycle cannot be $\mbr[n]$. In other words, the limit cycle cannot be globally asymptotically stable (GAS) in $\mbr[n]$.
	\end{lemma}
	\begin{proof}
		We prove by contradiction: Suppose that global convergence to the limit cycle $\mathcal{L}$ holds. Since the limit cycle is compact, it is an embedded submanifold in $\mbr[n]$ \cite[Proposition 5.21]{lee2015introduction}. So we can take a tubular neighborhood $\mathcal{O} \supseteq \mathcal{L}$ of the limit cycle \cite[Theorem 6.24]{lee2015introduction}. Then due to the asymptotic stability of the limit cycle, there exists a smaller neighborhood $\mathcal{U} \subseteq \mathcal{O}$ of the limit cycle such that every trajectory starting from $\mathcal{U}$ will remain within the tubular neighborhood $\mathcal{O}$ perpetually.  Since the limit cycle is compact, we can find a closed ball $\overline{\mathcal{B}} \subseteq \mbr[n]$ centered at $\bm{0} \in \mbr[n]$ sufficiently large such that the limit cycle lies in its interior (i.e., $\mathcal{L} \subseteq \mathcal{B}$). Due to the global convergence assumption, for any point $w \in \overline{\mathcal{B}}$, there exists a time instant $T_w > 0$ such that $\Psi^{T_w}(w) \in \mathcal{U}$, where $\Psi$ denotes the flow of the differential equation $\dot{x} = f(x)$. Due to the continuous dependence on initial conditions \cite[Theorem 3.5]{khalil2002nonlinear}, there exists an open set $\mathcal{V}_w \ni w$, such that $\Psi^{T_w}(\mathcal{V}_w) \subseteq \mathcal{U}$. Therefore, according to the uniqueness of solutions to the differential equation \cite[Theorem 3.1]{khalil2002nonlinear} and the asymptotic stability discussed before, we further have $\Psi^{t}(\mathcal{V}_w) \subseteq \mathcal{O}$ for all $t \ge T_w$. Thus, for every point $w \in \overline{\mathcal{B}}$, we can associate an open set $\mathcal{V}_w$ and a time instant $T_w$ as discussed before. Since $\mathcal{D} \defeq \{\mathcal{V}_w \subseteq \mbr[n] : w \in \overline{\mathcal{B}} \}$ is an open cover of the \emph{compact} ball $\overline{\mathcal{B}}$, there exists a finite number of points $w_i \in \overline{\mathcal{B}}, i=1,\dots,k$, and $\mathcal{V}_{w_i} \in \mathcal{D}$, such that $\bigcup_{i=1}^{k} \mathcal{V}_{w_i} \supseteq \overline{\mathcal{B}}$ \cite[Theorem 1.5.8]{tao2015analysis2}. Thus, we can take $T > \max_{i=1,\dots,k} \{T_{w_i}\}$, and therefore, we have $\Psi^{T}(\mathcal{B}) \subseteq \mathcal{O}$. 
		
		Let $r: \mathcal{O} \to \mathcal{L}$ be a retraction\footnote{The existence of $r$ is guaranteed by Proposition 6.25 in \cite{lee2015introduction}. }  of $\mathcal{O}$ onto $\mathcal{L}$; i.e., $r \circ i_{\mathcal{L}}=\identity_{\mathcal{L}}$, where $i_{\mathcal{L}}: \mathcal{L} \to \mathcal{O}$ is the inclusion map of $\mathcal{L}$ in $\mathcal{O}$ and $\identity$ is the identity map. Now let $i_{\mathcal{L}}': \mathcal{L} \to  \overline{\mathcal{B}}$ be another inclusion map, and note that for any $t \in \mathbb{R}$, $\Psi^{t}(\cdot)$ is a diffeomorphism of $\mathcal{L}$ \cite[p. 13]{chicone2006ordinary}. Then it is easy to check that $\identity_{\mathcal{L}}=\Psi^{-T} \circ r \circ \Psi^{T} \circ i_{\mathcal{L}}'$, where we view $\Psi^T$ as a map from $\overline{\mathcal{B}}$ to $\mathcal{O}$ and $\Psi^{-T}$ a map from $\mathcal{L}$ to $\mathcal{L}$. It is conventional to use $(\cdot)_*$ and $\pi_1(\cdot)$ to denote the homomorphism and the fundamental group respectively. Then 
		\begin{equation} \label{eq11}
		\begin{split}
		(\identity_{\mathcal{L}})_{*} &= (\Psi^{-T} \circ r \circ \Psi^{T} \circ i_{\mathcal{L}}')_* \\
		&= (\Psi^{-T})_* \circ (r)_* \circ (\Psi^{T})_* \circ (i_{\mathcal{L}}')_*,
		\end{split}
		\end{equation}
		where  $(\identity_{\mathcal{L}})_{*}: \pi_1(\mathcal{L}) \to \pi_1(\mathcal{L})$, $(i_{\mathcal{L}}')_*: \pi_1(\mathcal{L}) \to \pi_1(\overline{\mathcal{B}})$, $(\Psi^{T})_* : \pi_1(\overline{\mathcal{B}}) \to \pi_1(\mathcal{O})$, $r_{*}: \pi_1(\mathcal{O}) \to \pi_1(\mathcal{L})$ and $(\Psi^{-T})_*: \pi_1(\mathcal{L}) \to \pi_1(\mathcal{L})$ are the homomorphisms of fundamental groups induced by the corresponding maps \cite[Proposition A.64, A.65]{lee2015introduction}. Since $\overline{\mathcal{B}}$ is contractible and $\pi_1(\overline{\mathcal{B}}) \cong \{0\}$, where $\cong$ denotes the isomorphic relation, both $(i_{\mathcal{L}}')_*$ and $(\Psi^{T})_*$ are zero morphisms, and so is the composition $(\Psi^{-T})_* \circ (r)_* \circ (\Psi^{T})_* \circ (i_{\mathcal{L}}')_*$. But this contradicts with the left-hand side of \eqref{eq11}, where $(\identity_{\mathcal{L}})_{*}$ is the identity map (and an isomorphism) of $\pi_1(\mathcal{L}) \cong \mathbb{Z}$. The contradiction implies that global convergence is not possible.
	\end{proof}

	Based on Lemma \ref{lemma3_new}, we can prove Proposition \ref{prob2}.
	\begin{proof}[Proof of Proposition \ref{note_prop2new}]
		We consider the autonomous systems \eqref{eqode}. Without loss of generality, we assume that the flow of \eqref{eqode} is complete, since otherwise we can replace the vector field $\vf$ by $\vf / (1+\norm{\vf})$ without causing any difference of the topological properties of the flow that will be discussed in the sequel \cite[Proposition 1.14]{chicone2006ordinary}.
		
		Given $\alpha>0$, we define a neighborhood of the desired path $\setp$ by
		\begin{equation} \label{eq17}
		\mathcal{E}_\alpha = \{ \xi \in \mathbb{R}^n : \norm{e(\xi)} < \alpha \}.
		\end{equation}
		Therefore, the value of $\norm{e(\cdot)}$ encodes the \emph{distance} to the desired path in view of the definition of $\setp$ in \eqref{path2}. From Lemma \ref{lemma1}, we have $\transpose{N} \vf = \transpose{N}(\nabla_{\times}\phi - N K e) = -\transpose{N} N K e$. We define a Lyapunov function candidate 
		\begin{equation} \label{eq_lyapunov}
		V(e) = \frac{1}{2} \transpose{e} K e,
		\end{equation}
		and take the time derivative of it, obtaining
		\begin{equation} \label{eq15}
		\begin{split}
		\dot{V}(e) &= \frac{1}{2} (\transpose{\dot{e}} K e + \transpose{e} K \dot{e} ) \\
		&= \frac{1}{2} (\transpose{\vf} N K e + \transpose{e} K \transpose{N} \vf ) \\
		&= -\transpose{e} Q e = - \norm{N K e}^2 \le 0,
		\end{split}
		\end{equation}
		where the $(n-1) \times (n-1)$ matrix 
		\begin{equation} \label{eqQ}
		Q(\xi) = \transpose{K} \transpose{N}(\xi) N(\xi) K
		\end{equation}
		is positive semidefinite. Based on the LaSalle's invariance principle \cite[Theorem 4.4]{khalil2002nonlinear}, one can show that the desired path $\setp$ is the limit cycle of \eqref{eqode} by construction, and that $\setp$ is Lyapunov stable \cite{yao2018cdc}. The claim then easily follows from Lemma \ref{lemma3_new}.
	\end{proof}
	
	\section{Proofs for Section \ref{sec_extended}} \label{app_prj_dyn}
	\subsection{Proof of Lemma \ref{lemma2}}\label{appC1}
	\begin{proof}
		Due to the twice continuous differentiability of the transformation operator $G_l$, the corresponding Jacobian matrix function $\tangentmap G_l = \partial G_l / \partial x : \mbr[m] \to \mbr[m \times m]$ is locally Lipschitz continuous, where $x$ is the argument of $G_l$. Therefore, the product of the vector field $\vf$ and the Jacobian $\tangentmap G_l$ are also locally Lipschitz continuous. It follows that $\big( \xi(t), \trs{\xi}(t) \big) \in \mbr[2m]$, where $\trs{\xi}(t)=G_l \big( \xi(t) \big)$, is the unique solution to \eqref{eq6} \cite{chicone2006ordinary}. Recall that $l$ is the (global) Lipschitz constant of $G_l$. Fix $t$, then
		\begin{align*}
		\dist(\trs{\xi}(t), \trs{\mathcal{A}}) &= \inf\{ \norm{\trs{\xi}(t) - p} : p \in \trs{\mathcal{A}} \} \\
		&= \inf\{ \norm{G_l \big( \xi(t) \big) - G_l(q)} : q \in \mathcal{A} \} \\
		&\le \inf\{ l \norm{\xi(t) - q} : q \in \mathcal{A} \} \\
		&= l \cdot \dist(\xi(t), \mathcal{A}).
		\end{align*}
		Since $\xi(t)$ asymptotically converges to $\mathcal{A}$, we have $\dist(\xi(t), \mathcal{A}) \to 0$ as $t \to \infty$. In other words, for any $\epsilon>0$, there exists a $T>0$, such that for all $t \ge T$, $\dist(\xi(t), \mathcal{A}) < \epsilon / l$; hence $\dist(\trs{\xi}(t), \trs{\set{A}}) \le l \cdot \dist(\xi(t), \mathcal{A}) < \epsilon$. Therefore, $\dist(\trs{\xi}(t), \trs{\set{A}}) \to 0 $ as $t \to \infty$. Thus the transformed solution $\trs{\xi}(t)$ asymptotically converges to the transformed set $\trs{\set{A}}$.
	\end{proof}
	
	\subsection{Proof of Corollary \ref{corol_prj_dyn}}\label{appC2}
	\begin{proof}
		First consider the differential equation $\dot{\xi} = \vf(\xi)$. Using the same Lyapunov function candidate as \eqref{eq_lyapunov} and the argument in the proof of Proposition \ref{note_prop2new}, we have $\dot{V}(e) \le 0$. In addition, the norm of the first term of the scaled vector field $s \hat{\vf}(\xi)$ is $s \norm{\nabla_{\times} \phi}/ \norm{\vf}$, and it is obviously upper bounded in $\mbr[m]$. Since the new vector field $s \hat{\vf}(\xi)$ differs from the actual vector field $\vf(\xi)$ only by the magnitude of each vector, the two differential equations $\dot{\xi} = \vf(\xi)$ and $\dot{\xi} = s \hat{\vf}(\xi)$ have the same phase portrait in $\mathbb{R}^m \setminus \setc$ \cite[Proposition 1.14]{chicone2006ordinary}. Therefore, from the dichotomy convergence result proved in \cite[Theorem 3]{yao2018cdc}, the solution to $\dot{\xi} = s \hat{\vf}(\xi)$ will converge to either $\setp$ or $\setc$ for initial conditions $\xi(0) \in \mathbb{R}^m \setminus \setc$. 
		
		Note that if the maximal prolonged time is $t^* < \infty$, then the solution to $\dot{\xi} = s \hat{\vf}(\xi)$ must converge to the singular set $\setc$. This is shown by contradiction. Since $\normm{\dot{\xi}} = s < \infty$, $\xi^* \defeq \lim\limits_{t \to t^*} \xi(t) = \xi(0) + \int_{0}^{t^*} \dot{\xi}(t) dt$ exists. Suppose $\vf(\xi^*) \ne 0$, then we can define the solution at $t=t^*$, then it can be further prolonged to $t^*+\epsilon$ for some $\epsilon>0$, contradicting that $t^*$ is the maximal prolonged time. This shows that $\vf(\xi^*)=0$ and thus the solution converges to $\setc$.
		
		Finally, suppose $\xi(t)$ is the unique solution to the initial value problem $\dot{\xi}(t)= s \hat{\vf} \big( \xi(t) \big), \; \xi(0)=\xi_0$, then by Lemma \ref{lemma2}, $\big( \xi(t), \trs{\xi}(t) \big)$ is the solution to \eqref{eq7}. Therefore, the transformed trajectory $\trs{\xi}(t)$ asymptotically converges to the transformed desired path $\trs{\setp}$ as $t \rightarrow \infty$ or the transformed singular set $\trs{\setc}$ as $t \rightarrow t^*$.
	\end{proof}
	
	\section{Theoretical Guarantees for Features in Section \ref{sec_newgvf}} \label{app2}
	We use the set defined in \eqref{eq17} in the sequel, which corresponds to a neighborhood of the higher-dimensional desired path $\hgh{\setp}$ (rather than that of the physical desired path $\phy{\setp}$). Note that the higher-dimensional desired path $\hgh{\setp}$ is always unbounded and non-self-intersecting (i.e., homeomorphic to the real line), and thus an appealing result about the \emph{exponential convergence} of the norm of the path-following error $\norm{e(\cdot)}$ can be concluded from \cite[Theorem 2]{yao2020auto} with some modifications, where one of the conditions of the original theorem is automatically satisfied. Before we present the main results, we need a lemma regarding some algebraic calculations as follows.
	
	\begin{lemma} \label{lem_det}
		It holds that
		\[
		\det (\transpose{N} N) = \norm{\nabla_{\times} \phi}^2,
		\]
		where $N$ and $\nabla_{\times} \phi$ are defined in \eqref{eqvf2}.
	\end{lemma}
	\begin{proof}
		If $\nabla \phi_1, \dots, \nabla \phi_{n-1}$ are linearly dependent, then we have $\det (\transpose{N} N) = \norm{\nabla_{\times} \phi}^2=0$. Now suppose $\nabla \phi_1, \dots, \nabla \phi_{n-1}$ are linearly independent. In this case, it follows that $\norm{\nabla_{\times} \phi} \ne 0$. So we can define the normalized vector $v = \nabla_{\times} \phi / \norm{\nabla_{\times} \phi}$. Due to Lemma \ref{lemma1}, we have 
		\[
		\det (\transpose{N} N) = \det \left(\matr{\transpose{v} \\ \transpose{N}} \matr{v & N} \right) = \det \left(\matr{\transpose{v} \\ \transpose{N}} \right)^2.
		\]
		According to the definition of $\nabla_{\times} \phi$ in \eqref{eqvf2}, we have $\norm{\nabla_{\times} \phi} = \det \left(\matr{\transpose{v} \\ \transpose{N}} \right)$. Then it follows that $\det (\transpose{N} N) = \norm{\nabla_{\times} \phi}^2$.
	\end{proof}
	
	The result about the local and global exponential vanishing of the norm of the path-following error now follows.
	\begin{prop} \label{prop_expo}
		Define $\mathcal{E}_\alpha$ as in \eqref{eq17} for some $\alpha>0$, and let $\xi(t)$ be the solution to $\dot{\xi}(t)=\vf \big( \xi(t) \big)$, where $\vf: \mbr[n] \to \mbr[n]$ is the constructed guiding vector field on $\mbr[n]$ using the procedure in Section \ref{subsec_construct}. If $|f_i'(w)|$ is upper bounded in $\mathcal{E}_\alpha$ for $i=1,\dots,n-1$, then there exists a positive constant $\gamma' < \alpha$, such that the norm of the path-following error $\norm{e(t)}$ locally exponentially converges to $0$ as $t \to \infty$ when the initial condition $\xi(0) \in \mathcal{E}_{\gamma'}$. If $|f_i'(w)|$ is upper bounded globally, then the norm of the path-following error $\norm{e(t)}$ globally exponentially converges to $0$ as $t \to \infty$ .
	\end{prop}
	\begin{proof}
		The proof uses the same idea as in \cite[Theorem 2]{yao2020auto}. Using the construction of $\phi_i$ as in Section \ref{subsec_construct}, the norm of the ``tangential term'' $\norm{\nabla_{\times} \phi}$ is bounded if and only if $|f_i'(w)|$ is bounded. This condition guarantees that the solution is well defined for $t \to \infty$. In addition, since the last component of the term $\nabla_{\times}\phi$ is a nonzero constant, $\norm{\nabla_{\times}\phi}$ is automatically bounded away from zero globally in $\mbr[n]$. Namely, $\inf_{\xi \in \mbr[n]} \norm{\nabla_{\times}\phi(\xi)} > 0$. Due to Lemma \ref{lem_det}, we have $\lambda_1\dots\lambda_{n-1}=\det \big( Q(\xi) \big)=(k_1 \dots k_{n-1})^2 \norm{\nabla_{\times} \phi}^2$, where $\lambda_i$ is an eigenvalue of the matrix $Q(\xi)$ defined in \eqref{eqQ}. The previous reasoning and the boundedness of $|f_i'(w)|$ imply that $\Lambda'\defeq \inf_{\xi \in \mathcal{E}_{\alpha}} \lambda_{\mathrm{min}}\big( Q(\xi) \big)>0$. Using the function $V$ as in \eqref{eq_lyapunov}, we have $	\dot{V}\big( e(\xi) \big) \le -\Lambda' \norm{e(\xi)}^2 \le 2 \Lambda' V\big( e(\xi) \big) / k_{\mathrm{max}} $, further leading to $ ||e(\xi)|| \le c ||e_0|| \exp \left( {- \Lambda' t / k_{\mathrm{max}} } \right), $	where $e_0 = e\big( \xi(0) \big)$ and $c=\sqrt{k_{\mathrm{max}}/k_{\mathrm{min}}}$. Therefore, the error $\norm{e(\xi)}$ will locally exponentially converges to $0$ as $t$ approaches infinity.  If $|f_i'(w)|$ is upper bounded globally in $\mbr[n]$, then the solution to $\dot{\xi}(t)=\vf\big( \xi(t) \big)$ can be infinitely prolonged for every initial condition $\xi(0) \in \mbr[n]$, and thus the norm of the path-following error $\norm{e(t)}$ globally exponentially converges to $0$ as $t \to \infty$ .
	\end{proof}

	Due to Proposition \ref{prop_expo}, the dynamical system with the guiding vector field $\vf$ on a higher-dimensional space is robust against disturbance. To be more specific, suppose disturbance denoted by $d$ is added to the vector field; hence the perturbed system below is considered.
	\begin{equation} \label{eq_perturb}
	\dot{\xi}(t)=\vf\big( \xi(t) \big)+d(t),
	\end{equation}
	where $d: \mathbb{R}_{\ge 0} \to \mathbb{R}^n$, is a piecewise continuous and bounded function of time $t$ for all $t\ge0$. It can be shown that the path-following error dynamics
	\[
	\dot{e}(t) = \transpose{N}\big( \xi(t) \big) \big( \vf(\xi(t)) + d(t) \big)
	\]
	is locally input-to-state stable. This infers the following property of the path-following error.
	
	\begin{prop} \label{prop_iss}
		Consider the perturbed system \eqref{eq_perturb}. If $|f_i'(w)|$ is upper bounded in $\mathcal{E}_\alpha$ for $i=1,\dots,n-1$, then the error dynamics is locally input-to-state stable. Namely, there exists positive constants $\delta$ and $r$, such that if the initial condition $\xi(0) \in \mathcal{E}_\delta$ and the disturbance is uniformly bounded by $r$ (i.e., $\norm{d(t)} \le r$ for $t \ge 0$), then the norm of the path-following error $\norm{e(t)}$ is uniformly ultimately bounded. In addition, if the disturbance is vanishing; that is, $ \lim_{t \to \infty} \norm{d(t)} = 0$, then the norm of the path-following error $\norm{e(t)}$ is also vanishing as $t \to \infty$. 
	\end{prop}
	\begin{proof}
		The proof is similar to \cite[Theorem 3]{yao2020auto}, except that we consider a guiding vector field on $\mbr[n]$.
	\end{proof}
	
	\bibliographystyle{IEEEtran}
	\bibliography{ref}   

\begin{thebibliography}{10}
\providecommand{\url}[1]{#1}
\csname url@samestyle\endcsname
\providecommand{\newblock}{\relax}
\providecommand{\bibinfo}[2]{#2}
\providecommand{\BIBentrySTDinterwordspacing}{\spaceskip=0pt\relax}
\providecommand{\BIBentryALTinterwordstretchfactor}{4}
\providecommand{\BIBentryALTinterwordspacing}{\spaceskip=\fontdimen2\font plus
\BIBentryALTinterwordstretchfactor\fontdimen3\font minus
  \fontdimen4\font\relax}
\providecommand{\BIBforeignlanguage}[2]{{%
\expandafter\ifx\csname l@#1\endcsname\relax
\typeout{** WARNING: IEEEtran.bst: No hyphenation pattern has been}%
\typeout{** loaded for the language `#1'. Using the pattern for}%
\typeout{** the default language instead.}%
\else
\language=\csname l@#1\endcsname
\fi
#2}}
\providecommand{\BIBdecl}{\relax}
\BIBdecl

\bibitem{Siciliano:2007:SHR:1209344}
B.~Siciliano and O.~Khatib, \emph{Springer Handbook of Robotics}.\hskip 1em
  plus 0.5em minus 0.4em\relax Secaucus, NJ, USA: Springer-Verlag New York,
  Inc., 2007.

\bibitem{lacroix2016fleets}
S.~Lacroix, G.~Roberts, E.~Benard, M.~Bronz, F.~Burnet, E.~Bouhoubeiny, J.-P.
  Condomines, C.~Doll, G.~Hattenberger, F.~Lamraoui \emph{et~al.}, ``Fleets of
  enduring drones to probe atmospheric phenomena with clouds,'' in \emph{EGU
  General Assembly Conference Abstracts}, vol.~18, 2016.

\bibitem{Goncalves2010}
V.~M. Goncalves, L.~C.~A. Pimenta, C.~A. Maia, B.~C.~O. Dutra, and G.~A.~S.
  Pereira, ``Vector fields for robot navigation along time-varying curves in
  $n$-dimensions,'' \emph{IEEE Transactions on Robotics}, vol.~26, no.~4, pp.
  647--659, Aug 2010.

\bibitem{rezende2018robust}
A.~M. Rezende, V.~M. Gon{\c{c}}alves, G.~V. Raffo, and L.~C. Pimenta, ``Robust
  fixed-wing {UAV} guidance with circulating artificial vector fields,'' in
  \emph{2018 IEEE/RSJ International Conference on Intelligent Robots and
  Systems (IROS)}.\hskip 1em plus 0.5em minus 0.4em\relax IEEE, 2018, pp.
  5892--5899.

\bibitem{michalek2018vfo}
M.~M. Micha{\l}ek and T.~Gawron, ``{VFO} path following control with guarantees
  of positionally constrained transients for unicycle-like robots with
  constrained control input,'' \emph{Journal of Intelligent \& Robotic
  Systems}, vol.~89, no. 1-2, pp. 191--210, 2018.

\bibitem{hector2017}
H.~G. de~Marina, Z.~Sun, M.~Bronz, and G.~Hattenberger, ``Circular formation
  control of fixed-wing {UAV}s with constant speeds,'' in \emph{2017 IEEE/RSJ
  International Conference on Intelligent Robots and Systems (IROS)}, Sept
  2017, pp. 5298--5303.

\bibitem{liang2016combined}
Y.~Liang and Y.~Jia, ``Combined vector field approach for {2D} and {3D}
  arbitrary twice differentiable curved path following with constrained
  {UAV}s,'' \emph{Journal of Intelligent \& Robotic Systems}, vol.~83, no.~1,
  pp. 133--160, 2016.

\bibitem{KAPITANYUK20147342}
Y.~A. Kapitanyuk, S.~A. Chepinskiy, and A.~A. Kapitonov, ``Geometric path
  following control of a rigid body based on the stabilization of sets,''
  \emph{IFAC Proceedings Volumes}, vol.~47, no.~3, pp. 7342 -- 7347, 2014, 19th
  IFAC World Congress.

\bibitem{lakomy2017}
K.~{\L}akomy and M.~M. Micha{\l}ek, ``The {VFO} path-following kinematic
  controller for robotic vehicles moving in a {3D} space,'' in \emph{Robot
  Motion and Control (RoMoCo), 2017 11th International Workshop on}.\hskip 1em
  plus 0.5em minus 0.4em\relax IEEE, 2017, pp. 263--268.

\bibitem{michalek2010vector}
M.~Michalek and K.~Kozlowski, ``Vector-field-orientation feedback control
  method for a differentially driven vehicle,'' \emph{IEEE Transactions on
  Control Systems Technology}, vol.~18, no.~1, pp. 45--65, 2010.

\bibitem{caharija2016integral}
W.~Caharija, K.~Y. Pettersen, M.~Bibuli, P.~Calado, E.~Zereik, J.~Braga, J.~T.
  Gravdahl, A.~J. S{\o}rensen, M.~Milovanovi{\'c}, and G.~Bruzzone, ``Integral
  line-of-sight guidance and control of underactuated marine vehicles: Theory,
  simulations, and experiments,'' \emph{IEEE Transactions on Control Systems
  Technology}, vol.~24, no.~5, pp. 1623--1642, 2016.

\bibitem{borhaug2008integral}
E.~Borhaug, A.~Pavlov, and K.~Y. Pettersen, ``Integral {LOS} control for path
  following of underactuated marine surface vessels in the presence of constant
  ocean currents,'' in \emph{2008 47th IEEE Conference on Decision and
  Control}.\hskip 1em plus 0.5em minus 0.4em\relax IEEE, 2008, pp. 4984--4991.

\bibitem{yao2020auto}
W.~Yao and M.~Cao, ``Path following control in {3D} using a vector field,''
  \emph{Automatica}, vol. 117, p. 108957, 2020.

\bibitem{aguiar2008performance}
A.~P. Aguiar, J.~P. Hespanha, and P.~V. Kokotovi{\'c}, ``Performance
  limitations in reference tracking and path following for nonlinear systems,''
  \emph{Automatica}, vol.~44, no.~3, pp. 598--610, 2008.

\bibitem{seron1999feedback}
M.~M. Seron, J.~H. Braslavsky, P.~V. Kokotovic, and D.~Q. Mayne, ``Feedback
  limitations in nonlinear systems: From bode integrals to cheap control,''
  \emph{IEEE Transactions on Automatic Control}, vol.~44, no.~4, pp. 829--833,
  1999.

\bibitem{kapitanyuk2017guiding}
Y.~A. Kapitanyuk, A.~V. Proskurnikov, and M.~Cao, ``A guiding vector-field
  algorithm for path-following control of nonholonomic mobile robots,''
  \emph{IEEE Transactions on Control Systems Technology}, vol.~26, no.~4, pp.
  1372--1385, 2017.

\bibitem{lawrence2008lyapunov}
D.~A. Lawrence, E.~W. Frew, and W.~J. Pisano, ``Lyapunov vector fields for
  autonomous unmanned aircraft flight control,'' \emph{Journal of Guidance,
  Control, and Dynamics}, vol.~31, no.~5, pp. 1220--1229, 2008.

\bibitem{Sujit2014}
P.~B. Sujit, S.~Saripalli, and J.~B. Sousa, ``Unmanned aerial vehicle path
  following: A survey and analysis of algorithms for fixed-wing unmanned aerial
  vehicless,'' \emph{IEEE Control Systems}, vol.~34, no.~1, pp. 42--59, Feb
  2014.

\bibitem{phillips2004mechanics}
W.~F. Phillips, \emph{Mechanics of flight}.\hskip 1em plus 0.5em minus
  0.4em\relax John Wiley \& Sons, 2004.

\bibitem{rubi2019survey}
B.~Rub{\'\i}, R.~P{\'e}rez, and B.~Morcego, ``A survey of path following
  control strategies for {UAV}s focused on quadrotors,'' \emph{Journal of
  Intelligent \& Robotic Systems}, pp. 1--25, 2019.

\bibitem{caharija2015comparison}
W.~Caharija, K.~Y. Pettersen, P.~Calado, and J.~Braga, ``A comparison between
  the {ILOS} guidance and the vector field guidance,''
  \emph{IFAC-PapersOnLine}, vol.~48, no.~16, pp. 89--94, 2015.

\bibitem{yao2018cdc}
W.~Yao, Y.~A. Kapitanyuk, and M.~Cao, ``Robotic path following in {3D} using a
  guiding vector field,'' in \emph{IEEE Conference on Decision and Control},
  2018, pp. 4475--4480.

\bibitem{de2017guidance}
H.~G. De~Marina, Y.~A. Kapitanyuk, M.~Bronz, G.~Hattenberger, and M.~Cao,
  ``Guidance algorithm for smooth trajectory tracking of a fixed wing {UAV}
  flying in wind flows,'' in \emph{2017 IEEE International Conference on
  Robotics and Automation (ICRA)}.\hskip 1em plus 0.5em minus 0.4em\relax IEEE,
  2017, pp. 5740--5745.

\bibitem{lee2015introduction}
J.~Lee, \emph{Introduction to Smooth Manifolds: Second Edition}, ser. Graduate
  texts in mathematics.\hskip 1em plus 0.5em minus 0.4em\relax Springer, 2015.

\bibitem{nelson2007vector}
D.~R. Nelson, D.~B. Barber, T.~W. McLain, and R.~W. Beard, ``Vector field path
  following for miniature air vehicles,'' \emph{IEEE Transactions on Robotics},
  vol.~23, no.~3, pp. 519--529, 2007.

\bibitem{fossen2003line}
T.~I. Fossen, M.~Breivik, and R.~Skjetne, ``Line-of-sight path following of
  underactuated marine craft,'' \emph{IFAC Proceedings Volumes}, vol.~36,
  no.~21, pp. 211--216, 2003.

\bibitem{Zhu2013}
S.~Zhu, D.~Wang, and C.~B. Low, ``Ground target tracking using {UAV} with input
  constraints,'' \emph{Journal of Intelligent {\&} Robotic Systems}, vol.~69,
  no.~1, pp. 417--429, Jan 2013.

\bibitem{soetanto2003adaptive}
D.~Soetanto, L.~Lapierre, and A.~Pascoal, ``Adaptive, non-singular
  path-following control of dynamic wheeled robots,'' in \emph{Decision and
  Control, 2003. Proceedings. 42nd IEEE Conference on}, vol.~2.\hskip 1em plus
  0.5em minus 0.4em\relax IEEE, 2003, pp. 1765--1770.

\bibitem{sgorbissa20103D}
A.~Sgorbissa and R.~Zaccaria, ``{3D} path following with no bounds on the path
  curvature through surface intersection,'' in \emph{Intelligent Robots and
  Systems (IROS), 2010 IEEE/RSJ International Conference on}.\hskip 1em plus
  0.5em minus 0.4em\relax IEEE, 2010, pp. 4029--4035.

\bibitem{do2016differential}
M.~P. Do~Carmo, \emph{Differential Geometry of Curves and Surfaces: Revised and
  Updated Second Edition}.\hskip 1em plus 0.5em minus 0.4em\relax Courier Dover
  Publications, 2016.

\bibitem{lee2010topologicalmanifolds}
J.~Lee, \emph{Introduction to topological manifolds}.\hskip 1em plus 0.5em
  minus 0.4em\relax Springer Science \& Business Media, 2010, vol. 202.

\bibitem{mondada2009puck}
F.~Mondada, M.~Bonani, X.~Raemy, J.~Pugh, C.~Cianci, A.~Klaptocz, S.~Magnenat,
  J.-C. Zufferey, D.~Floreano, and A.~Martinoli, ``The e-puck, a robot designed
  for education in engineering,'' in \emph{Proceedings of the 9th conference on
  autonomous robot systems and competitions}, vol.~1, no.
  LIS-CONF-2009-004.\hskip 1em plus 0.5em minus 0.4em\relax IPCB: Instituto
  Polit{\'e}cnico de Castelo Branco, 2009, pp. 59--65.

\bibitem{yao2020mobile2}
W.~Yao, H.~G. de~Marina, and M.~Cao, ``Vector field guided path following
  control: Singularity elimination and global convergence,'' in \emph{2020 IEEE
  59th Conference on Decision and Control (CDC)}, 2020, to appear, arXiv
  preprint arXiv:2003.10012.

\bibitem{khalil2002nonlinear}
H.~Khalil, \emph{Nonlinear Systems}, 3rd~ed.\hskip 1em plus 0.5em minus
  0.4em\relax Prentice Hall, 2002.

\bibitem{yao2021topo}
\BIBentryALTinterwordspacing
W.~Yao, B.~Lin, B.~D.~O. Anderson, and M.~Cao, ``Topological analysis of
  vector-field guided path following on manifolds,'' 2021, submitted. [Online].
  Available: \url{http://tiny.cc/yao_topo}
\BIBentrySTDinterwordspacing

\bibitem{consolini2010path}
L.~Consolini, M.~Maggiore, C.~Nielsen, and M.~Tosques, ``Path following for the
  {PVTOL} aircraft,'' \emph{Automatica}, vol.~46, no.~8, pp. 1284--1296, 2010.

\bibitem{morro2011path}
A.~Morro, A.~Sgorbissa, and R.~Zaccaria, ``Path following for unicycle robots
  with an arbitrary path curvature,'' \emph{IEEE Transactions on Robotics},
  vol.~27, no.~5, pp. 1016--1023, 2011.

\bibitem{do2015global}
K.~D. Do, ``Global output-feedback path-following control of unicycle-type
  mobile robots: A level curve approach,'' \emph{Robotics and Autonomous
  Systems}, vol.~74, pp. 229--242, 2015.

\bibitem{chen2011curve}
Y.-Y. Chen and Y.-P. Tian, ``A curve extension design for coordinated path
  following control of unicycles along given convex loops,''
  \emph{International Journal of Control}, vol.~84, no.~10, pp. 1729--1745,
  2011.

\bibitem{galbis2012vector}
A.~Galbis and M.~Maestre, \emph{Vector analysis versus vector calculus}.\hskip
  1em plus 0.5em minus 0.4em\relax Springer Science \& Business Media, 2012.

\bibitem{yao2019integrated}
W.~Yao, B.~Lin, and M.~Cao, ``Integrated path following and collision avoidance
  using a composite vector field,'' in \emph{2019 IEEE 58th Conference on
  Decision and Control (CDC)}, 2019, pp. 250--255.

\bibitem{kapitanyuk2017guiding2}
Y.~A. Kapitanyuk, H.~G. de~Marina, A.~V. Proskurnikov, and M.~Cao, ``Guiding
  vector field algorithm for a moving path following problem,''
  \emph{IFAC-PapersOnLine}, vol.~50, no.~1, pp. 6983--6988, 2017.

\bibitem{chicone2006ordinary}
C.~Chicone, \emph{Ordinary differential equations with applications}.\hskip 1em
  plus 0.5em minus 0.4em\relax Springer Science \& Business Media, 2006,
  vol.~34.

\bibitem{tao2015analysis2}
T.~Tao, \emph{Analysis II, Texts and Readings in Mathematics, vol. 38}.\hskip
  1em plus 0.5em minus 0.4em\relax Springer, New York, 2015.

\bibitem{yao2021multi}
\BIBentryALTinterwordspacing
W.~Yao, H.~G. de~Marina, Z.~Sun, and M.~Cao, ``Distributed coordinated path
  following using guiding vector fields,'' in \emph{IEEE International
  Conference on Robotics and Automation (ICRA)}, 2021, submitted, full version.
  [Online]. Available: \url{http://tiny.cc/yao_icra21}
\BIBentrySTDinterwordspacing

\bibitem{gati2013open}
B.~Gati, ``Open source autopilot for academic research-the paparazzi system,''
  in \emph{2013 American Control Conference}.\hskip 1em plus 0.5em minus
  0.4em\relax IEEE, 2013, pp. 1478--1481.

\bibitem{siciliano2010robotics}
B.~Siciliano, L.~Sciavicco, L.~Villani, and G.~Oriolo, \emph{Robotics:
  modelling, planning and control}.\hskip 1em plus 0.5em minus 0.4em\relax
  Springer Science \& Business Media, 2010.

\bibitem{yao2021dichotomy}
\BIBentryALTinterwordspacing
W.~Yao, B.~Lin, B.~D.~O. Anderson, and M.~Cao, ``Refining dichotomy convergence
  in vector-field guided path following control,'' in \emph{European Control
  Conference (ECC)}, 2021, submitted. [Online]. Available:
  \url{http://tiny.cc/yao_ecc21}
\BIBentrySTDinterwordspacing

\bibitem{giaquinta2010mathematical}
M.~Giaquinta and G.~Modica, \emph{Mathematical analysis: an introduction to
  functions of several variables}.\hskip 1em plus 0.5em minus 0.4em\relax
  Springer Science \& Business Media, 2010.

\end{thebibliography}
	
	\begin{IEEEbiography}[{\includegraphics[width=1in,height=1.25in,clip,keepaspectratio]{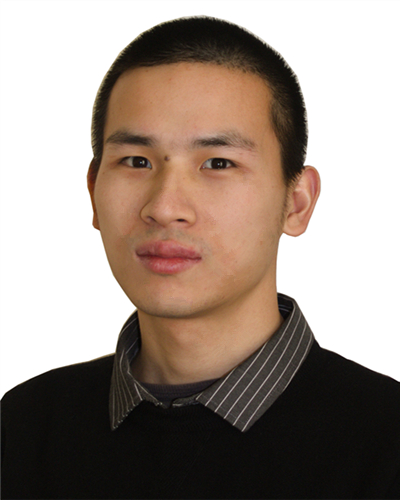}}]{Weijia Yao} has since 2017 been a Ph.D. student with the Engineering and Technology Institute (ENTEG), University of Groningen, the Netherlands. He received the Bachelor and Master degrees from National University of Defense Technology, China, in 2015 and 2017 respectively, both in control science and engineering. He received the award of Outstanding Master Degree Dissertation of Hunan province, China, in 2020.  His research interests include nonlinear systems and control, robotics and multi-agent systems. 
	\end{IEEEbiography}
	
	\begin{IEEEbiography}[{\includegraphics[width=1in,height=1.25in,clip,keepaspectratio]{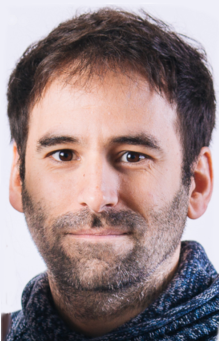}}]{H\'ector Garc\'ia de Marina} received the Ph.D. degree in systems and control from the University of Groningen, the Netherlands, in 2016. He was a post-doctoral research associate with the Ecole Nationale de l'Aviation Civile, Toulouse, France, and an assistant professor at the Unmanned Aerial Systems Center in the University of Southern Denmark. He is currently a researcher in the Department of Computer Architecture and Automatic Control at the Faculty of Physics, Universidad Complutense de Madrid, Spain. His current research interests include multi-agent systems and the design of guidance navigation and control systems for autonomous vehicles.
	\end{IEEEbiography}

	\begin{IEEEbiography}[{\includegraphics[width=1in,height=1.25in,clip,keepaspectratio]{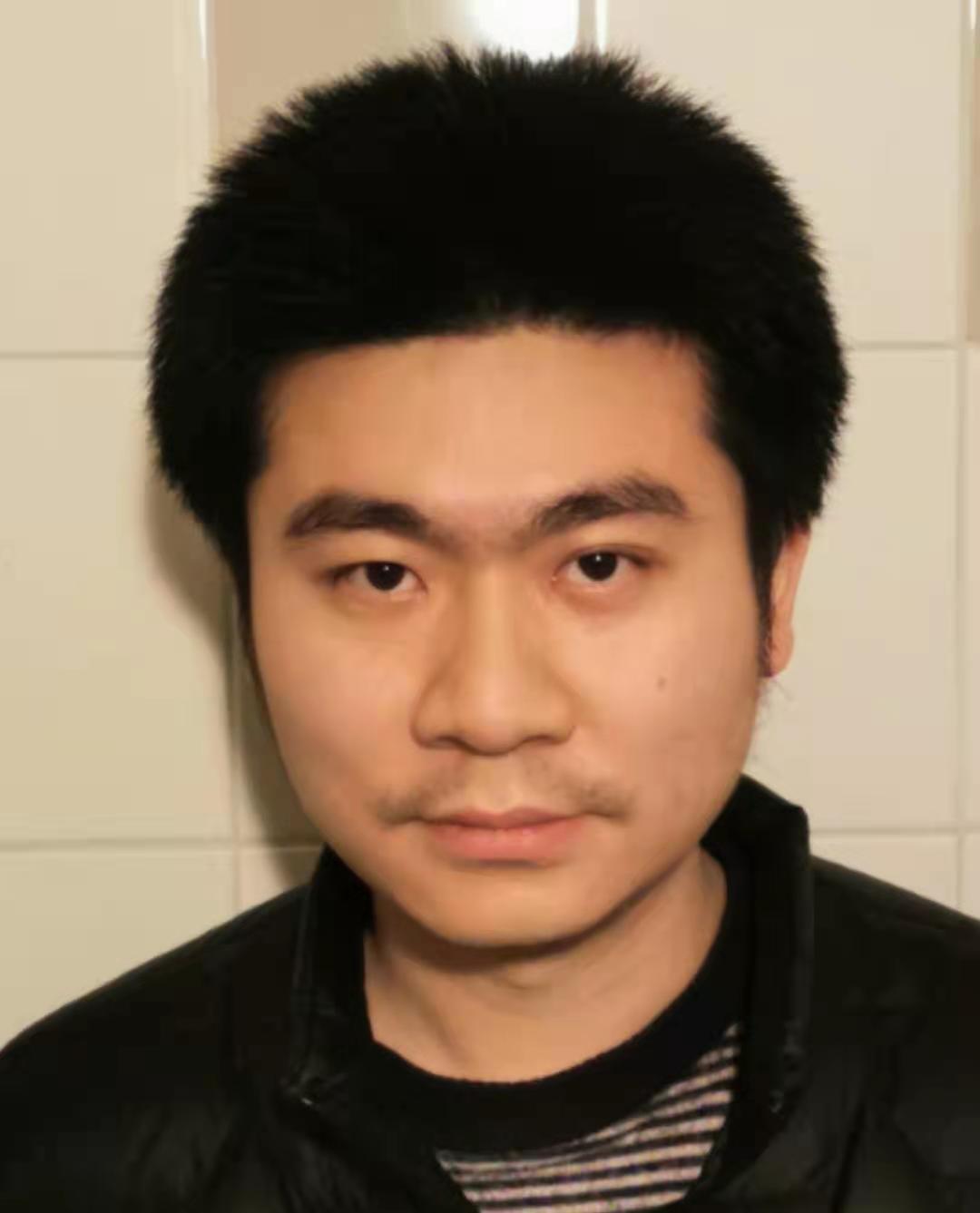}}]{Bohuan Lin} received his bachelor's degree from Sun Yat-sen University (mainland, China) for his first major in Theoretical and Applied Mechanics in 2013, and completed a second major in mathematics in 2014. He then completed his master's degree at the School of Mathematics of the same university in 2016 with his work on a certain PDE. Now he is working on a PhD project on (Hamiltonian/non-Hamiltonian) integrable systems at the Bernoulli Institute of the University of Groningen, the Netherlands.
	\end{IEEEbiography}
	
	\begin{IEEEbiography}[{\includegraphics[width=1in,height=1.25in,clip,keepaspectratio]{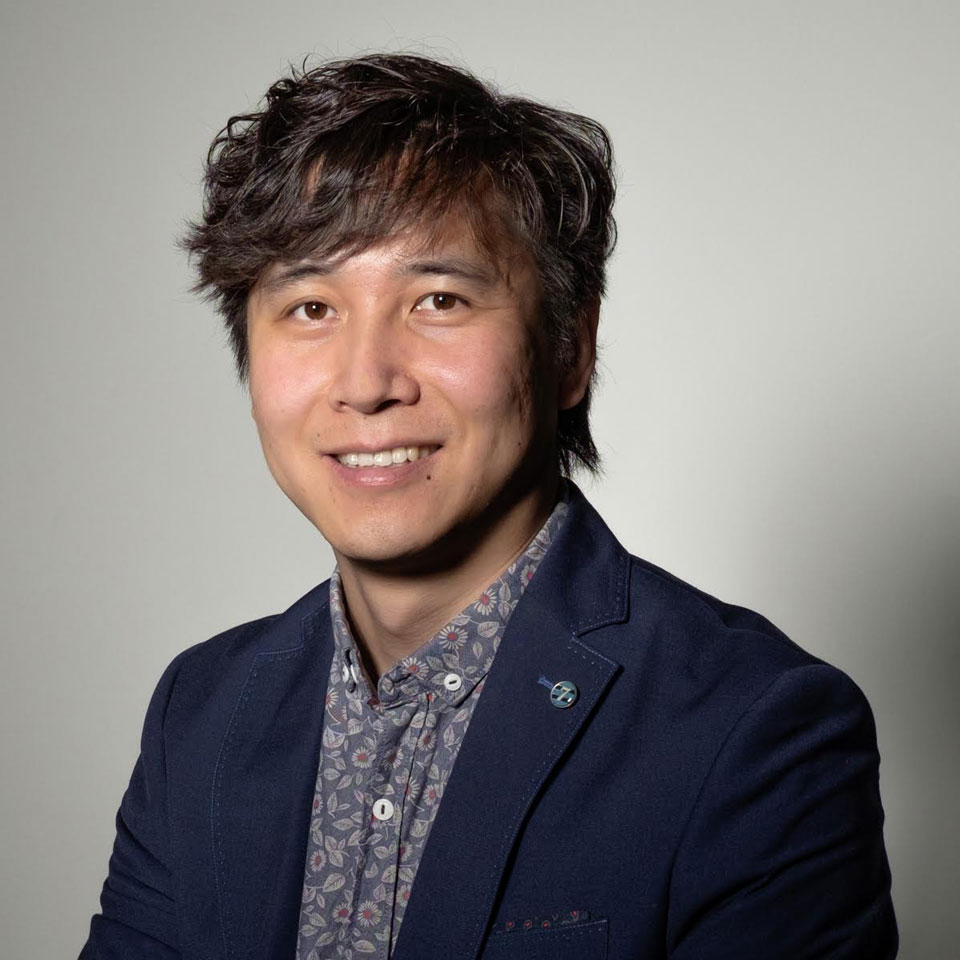}}]{Ming Cao}  has since 2016 been a professor of systems and control with the Engineering and Technology Institute (ENTEG) at the University of Groningen, the Netherlands, where he started as a tenure-track Assistant Professor in 2008. He received the Bachelor degree in 1999 and the Master degree in 2002 from Tsinghua University, Beijing, China, and the Ph.D. degree in 2007 from Yale University, New Haven, CT, USA, all in Electrical Engineering. From September 2007 to August 2008, he was a Postdoctoral Research Associate with the Department of Mechanical and Aerospace Engineering at Princeton University, Princeton, NJ, USA. He worked as a research intern during the summer of 2006 with the Mathematical Sciences Department at the IBM T. J. Watson Research Center, NY, USA. He is the 2017 and inaugural recipient of the Manfred Thoma medal from the International Federation of Automatic Control (IFAC) and the 2016 recipient of the European Control Award sponsored by the European Control Association (EUCA). He is a Senior Editor for Systems and Control Letters, and an Associate Editor for IEEE Transactions on Automatic Control, IEEE Transactions on Circuits and Systems and IEEE Circuits and Systems Magazine. He is a vice chair of the IFAC Technical Committee on Large-Scale Complex Systems. His research interests include autonomous agents and multi-agent systems, complex networks and decision-making processes.   
	\end{IEEEbiography}
	\vfill
	
\end{document}